\documentclass{article}

\usepackage[final]{nips_2017}
\usepackage{booktabs}
\usepackage{stackrel}
\usepackage{times} 
\usepackage[geometry]{ifsym} 
\usepackage[utf8]{inputenc} 
\usepackage{amsmath,amssymb,amsthm} 
\usepackage{graphicx} 
\usepackage{subfigure} 
\usepackage{booktabs} 
\usepackage{multirow} 
\usepackage{multicol}
\usepackage[dvipsnames]{xcolor}
\usepackage{url}
\usepackage{caption}
\usepackage{paralist}
\usepackage{algpseudocode,algorithm,algorithmicx}
\usepackage{titlesec} 
\usepackage{xspace}
\usepackage{color}
\usepackage{listings}
\usepackage[section]{placeins} 

\definecolor{mygreen}{rgb}{0,0.6,0}
\definecolor{mygray}{rgb}{0.5,0.5,0.5}
\definecolor{mymauve}{rgb}{0.58,0,0.82}

\graphicspath{figures}

\usepackage[utf8]{inputenc} 
\usepackage[T1]{fontenc}    
\usepackage{hyperref}       
\usepackage{amsfonts}       
\usepackage{nicefrac}       
\usepackage{microtype}      

\usepackage{centernot}
\makeatletter
\newcommand*{\cnternot}{%
  \mathpalette\@cnternot
}
\def\@cnternot#1#2{%
  \mathrel{%
    \rlap{%
      \settowidth\dimen@{$\m@th#1{#2}$}%
      \kern.5\dimen@
      \settowidth\dimen@{$\m@th#1=$}%
      \kern-.5\dimen@
      $\m@th#1\not$%
    }%
    {#2}%
  }%
}
\makeatother

\newcommand{\ttdim}{\texttt{dim}{}}
\newcommand{\ttsamp}{\texttt{n\_samples}{}}

\DeclareMathOperator{\independent}{\perp\mkern-9.5mu\perp}
\DeclareMathOperator{\notindependent}{\centernot{\independent}}

\title{Fast Conditional Independence Test for Vector Variables with Large Sample Sizes}

\author{
  Krzysztof Chalupka\\
  Computation and Neural Systems\\
  California Institute of Technology\\
  Pasadena, CA 91106 \\
  \And
  Pietro Perona \\
  Electrical Engineering \\
  California Institute of Technology\\
  Pasadena, CA 91106 \\
  \And 
  Frederick Eberhardt \\
  Humanities and Social Sciences\\
  California Institute of Technology\\
  Pasadena, CA 91106\\
  \texttt{fde@caltech.edu}
}

\begin{document}
\maketitle

\begin{abstract} We present and evaluate the Fast (conditional) Independence Test (FIT) -- a nonparametric conditional independence test. The test is based on the idea that when $P(X \mid Y, Z) = P(X \mid Y)$, $Z$ is not useful as a feature to predict $X$, as long as $Y$ is also a regressor. On the contrary, if $P(X \mid Y, Z) \neq P(X \mid Y)$, $Z$ might improve prediction results. FIT applies to thousand-dimensional random variables with a hundred thousand samples in a fraction of the time required by alternative methods. We provide an extensive evaluation that compares FIT to six extant nonparametric independence tests. The evaluation shows that FIT has low probability of making both Type I and Type II errors compared to other tests, especially as the number of available samples grows. Our implementation of FIT is publicly available\footnote{\url{https://github.com/kjchalup/fcit}}.
\end{abstract}

\section{Introduction}
Two random variables $X$ and $Y$ are conditionally independent given a third variable $Z$ if and only if $P(X, Y \mid Z) = P(X \mid Z) P(Y \mid Z)$. We denote this relationship by $X \independent Y \mid Z$, and its negation, the case of conditional dependence, by $X \notindependent Y \mid Z$. This article develops and evaluates the Fast (conditional) Independence Test (FIT), a nonparametric conditional or unconditional independence test. Given a finite sample from the joint distribution $P(X, Y, Z)$, FIT returns the p-value under the null hypothesis that $X\independent Y \mid Z$. FIT applies to datasets with large sample sizes on scalar- or vector-valued variables, and returns in short time. Extensive empirical comparison with existing alternatives shows that FIT is currently the only conditional independence test to achieve this goal.

\subsection{Motivation}
\label{sec:motivation}
Independence tests are ubiquitous in scientific inference. They are the main work horse of classical hypothesis testing (see~\citet{fisher1992}, Chapter 21), which underlies most experimental and many observational techniques of data analysis. A conditional independence test (CIT) extends this idea by testing for independence between two variables given the value of a third, or more generally, given the values of a set of further variables~\citep{dawid1979}.
Probabilistic dependence -- conditional or not -- is often taken as an indicator of informational exchange (see e.g.~\citet{cover2012}, Chapter 2), causal connection~\citep{dawid1979,spirtes_causation_2000} or simply as tool for prediction and diagnosis~\citep{koller1996}. Consequently, knowledge of the (conditional) independence relations among a set of variables provides some of the most basic indicators for scientific relations of interest.

Our motivation comes from the challenge of causal discovery. Many extant causal discovery algorithms use conditional independence tests to infer the underlying causal structure over a set of variables from observational data~\citep{pearl_causality_2000,spirtes_causation_2000,hoyer2009}. The most general of these methods do not make any specific assumptions about the functional form or parameterization of the underlying causal system. Discovery of independence structure from data, rather than estimation of any specific parameters, allows these methods to establish causal structure. However, such generality can only be maintained if the CITs used to determine the independence structure in the first place are themselves non-parametric. That is, the CITs have to check for general dependence relations among the variables, and not just, say, for a non-zero (partial) correlation.

A further motivation derives from the need for CITs that are both fast to compute and flexible in the variables they can handle. Generally, causal discovery algorithms have to perform a number of tests that is at least polynomial, if not exponential, in the number of variables~\citep{spirtes_causation_2000}. For genetic datasets where thousands of genes are measured, or neuroscience data containing tens of thousands of voxels, this implies millions of independence tests. Moreover, for example, in social network data, the variables may be categorical or continuous, and scalar or vector valued. These facts indicate a clear need for non-parametric tests that are fast and flexible with regard to the nature of the variables being tested.

While our motivation derives from the domain of causal discovery, the test is completely domain general and can be applied to any area in which (conditional) independence tests are used.

\subsection{Previous work}
\label{sec:alternatives}
For continuous variables, extant non-parametric CITs have largely focused on kernel methods~\citep{scholkopf2002}. The methods differ in how the null-distribution is estimated, which distance measure is used in the reproducing kernel Hilbert space (RKHS) or which test statistic is applied. Here we consider six methods we are aware of for comparison:

The Conditional Hilbert-Schmidt Independence Criterion (CHSIC) \citep{fukumizu2008} maps the data into a RKHS. It determines independence using the normalized cross-covariance operator, which can be thought of as an extension of standard correlations to higher order moments. 

In contrast, the Kernel-based Conditional Independence test (KCIT) \citep{zhang_kernel-based_2012} derives a test statistic based on the traces of the kernel matrices and approximates the null-distribution using a Gamma distribution. This avoids the permutations required for CHSIC, but comes at the expense of significant matrix operations. 

The Kernel Conditional Independence Permutation test (KCIPT) \citep{doran_permutation-based_2014}, in many ways a variant of CHSIC, replaces the random sampling of permutations for the null-distribution with a specifically learned permutation that satisfies, among other criteria, that it is representative of conditional independence. KCIPT also replaces the normalized cross-covariance operator with the maximum mean discrepancy (MMD) test statistic developed by \citet{gretton_kernel_2012}. 

Kernel-based tests do not scale well to large sample sizes due to the matrix operations required to kernelize the data and embed it in the RKHS. To address this limitation, the Randomized Conditional Independence test (RCIT)~\citep{strobl_approximate_2017}, uses a fast Fourier transform to speed up the matrix operations.
For similar reasons, the Conditional Correlation Independence test (CCI) \citep{ramsey_scalable_2014} determines independence of $X$ and $Y$ by only checking whether $cov(f(X), g(Y)) = 0$ for different choices of $f, g$ taken from an (in practice, truncated) set of basis functions. For conditional independence of $X \independent Y \mid Z$, the same test is performed on the non-parametric residuals of $X$ and $Y$ each regressed on $Z$ using a uniform kernel. 

In the case of categorical data, two variables are independent conditional on a third categorical variable $Z$ just in case they are independent given any specific categorical value of $Z=z$. Consequently, the simplest way to handle categorical data is to repeat whatever unconditional test is available for each subset of the data corresponding to a fixed value of $Z=z$. Unless some further structure among the categories (such as a particular order) holds important information about the probabilistic relations between variables, there is nothing further for a conditional independence test to exploit. Consequently, tests conditioning on categorical variables are in the general case no different than unconditional tests, and inevitably sample intensive. However, a particular -- and common -- challenge arises when the conditioning variable $Z$ is continuous and at least one of $X$ or $Y$ is categorical. In this case we have a "hybrid" situation. The methods mentioned above do not apply due to the categorical nature of $X$ or $Y$,  and for a continuous conditioning variable $Z$, by definition one cannot explicitly check conditional independence by checking independence for each value of $Z$. 

The development of the above tests has enabled the construction of genuinely non-parametric methods of discovery, but has left several questions of practical importance open. First, it remains unclear how to choose between the tests given that there is no available systematic comparison. Second, the evaluation of tests has focused on a small set of specialized datasets. Third, there has been no evaluation of how these tests perform for high dimensional variables and large sample sizes. 

In our evaluation in Sec.~\ref{sec:results} we have attempted to faithfully implement all the above tests\footnote{For all but the CCI test, we wrote Python wrappers around existing Matlab or R implementations written by the original authors. We implemented CCI from scratch in Numpy.} and systematically evaluated them on all the previous data sets, in each case reporting the actual behavior of the resulting $p$-values. In addition, we vastly extended the range of parameters that were explored to generate some of the existing datasets, and added datasets that explore the hybrid case with categorical and continuous variables described above.

\subsection{Contributions}
We address the challenge of testing for (conditional) independence when the variables are high-dimensional and sample sizes are high, but when the relation among the variables cannot be well described by a parametric model. Moreover, in order to enable applications for which it is necessary to run tens of thousands of conditional independence tests
-- for example, establishing the causal graph over hundreds of variables -- the test has to run in a short amount of time.

To our knowledge, none of the existing CITs applies with reasonable runtime in such situations. The main contributions of this article are as follows:

\begin{compactenum}
\item The Fast Independence Test (FIT), which can process hundred-dimensional conditional independence queries with $10^5$ samples in less than 60s.
\item An extensive evaluation and comparison of FIT and alternatives (CHSIC, KCIT, KCIPT, RCIT, CCI) on a wide range of datasets.
\item A parallelized Python implementation of FIT, easily installable through Python Package Index\footnote{To install FIT on Linux, type \texttt{pip install fcit} in the terminal window.}.
\end{compactenum}

\section{Fast (conditional) Independence Test (FIT)}
The intuition behind FIT is that if $X \notindependent Y \mid Z$, then prediction of $Y$ using both $X$ and $Z$ as covariates should be more accurate than prediction of $Y$ when only $Z$ is used as the covariate. On the other hand, if $X \independent Y \mid Z$, then the accuracy of the prediction of $Y$ should not change whether just $Z$, or $X$ \emph{and} $Z$ are used as covariates. While this is not always true (see Sec~\ref{sec:discussion_failure}), it is a relatively weak assumption to make compared to, for example, assuming a specific parametric form of the involved distributions or linearity of the functional relationships. Alg.~\ref{alg:fit} contains pseudocode for the implementation of FIT.

Our implementation uses decision tree regression (DTR,~\citet{breiman_classification_1984})\footnote{DTR is a fast multi-output regression algorithm. It splits the input space in two on each decision tree node until reaching a leaf. A leaf assigns a constant value to each output dimension. During training, the \texttt{min\_samples\_split} hyperparameter turns a node into a leaf when the number of samples in the node falls below the threshold.} to predict $Y$ using both $X, Z$, and also using $Z$ only. he rationale behind this choice and the alternatives are discussed in Sec.~\ref{sec:discussion_ml}.

We measure the accuracy of the prediction in terms of the mean squared error (MSE). Central to our approach is the simplifying assumption that $X \independent Y \mid Z$ if and only if the mean squared error (MSE) of the algorithm trained using both $X$ and $Z$ is not smaller than the MSE of the algorithm trained using $Z$ only.

\subsection{Obtaining the p-value}
Our implementation runs DTR \texttt{n\_perm} times to learn both the function $X, Z \mapsto Y$ as well as $Z \mapsto Y$. The train/test (or in-sample/out-of-sample) split is resampled each of the \texttt{n\_perm} training repetitions. We store the \texttt{n\_perm} MSEs resulting from regressing $Y$ on $X, Z$ in a list $\texttt{mses\_x}$, and the \texttt{n\_perm} MSEs resulting from regressing $Y$ on $Z$ in a list \texttt{mses\_nox}. 

The remaining challenge is to determine the p-value of the null hypothesis that the first list contains on average values equal to or larger than the second list. To do this, we run a one-tailed t-test~\citet{student1908}, with the null hypothesis that the corresponding values in the two arrays are on average equal. A small p-value rejects the null hypothesis, meaning that \texttt{mses\_x} are significantly smaller than \texttt{mses\_nox} or that $X \notindependent Y \mid Z$. As an alternative to using the t-test (which assumes the distribution of MSEs under the null hypothesis is Gaussian), we experimented with using bootstrap to estimate the null distribution~\citep{efron1992}. Since the results were nearly identical across our evaluation tasks, for simplicity of presentation we use the t-test in the final method.

\subsection{Speed of the Implementation}
Two details of the implementation are worth mentioning:

\begin{compactenum}
\item Both the cross-validation procedure (Lines~\ref{lin:cv1},~\ref{lin:cv2}) and the training loop (Lines~\ref{lin:tr1}-~\ref{lin:z3}) are parallelized to automatically take up all the available CPUs. This provides a multiplicative speedup for larger datasets.
\item Using our algorithm as an unconditional independence test (testing whether $X \independent Y$) requires only a simple modification, which our implementation uses automatically when $Z$ is not given. Instead of training using only $Z$ as input (Lines~\ref{lin:cv2} and~\ref{lin:z2},~\ref{lin:z3}) we train using $X$ permuted randomly across the sampling dimension.
\end{compactenum}

Asymptotically the algorithm's speed scales as $\mathcal{O}(n\_samples^2 \times \log{(n\_samples)} \times n\_features \times \frac{1}{n\_cpus})$. Sec.~\ref{sec:results} shows actual runtimes on an 8 Intel(R) Core(TM) i7-6700K CPU \@ 4.00GHz machine. The runtimes are bounded below by about 1s due to parallelization overhead, and grow to about 100s for a 1000-dimensional dataset with 100K samples.

\begin{algorithm}
\caption{Fast (conditional) Independence Test}
\label{alg:fit}
\begin{lstlisting}[language=python,escapechar=&]
def cross_validate(covarite, regressand):
    # Find the best decision tree hyperparameters for 
    # regressing `regressand` on `covariate`. 
    
    # Return a trainable decision tree with the best hyperparameters.
    
    
def fit_test(x, y, z, n_perm=8, frac_test=.1):
    # Store total sample size and test set size.
    n_samples = x.shape[0]
    n_test = floor(frac_test * n_samples)
    
    # Find best decision tree parameters for learning y = f(x, z).&\label{lin:cv1}&
    best_tree_x = cross_validate(concat(x, z), y)
    
    # Find the best decision tree parameters for learning y = f(z).
    best_tree_nox = cross_validate(z, y)&\label{lin:cv2}&

    # Obtain `n_perm` MSEs on random train-test splits.
    mses_x = list()
    mses_nox = list()
    
    # In our implementation, this loop is parallelized.&\label{lin:tr1}&
    for perm_id in range(n_perm):
    	perm_ids = random.permutation(n_samples)
        x_test, x_train = x[perm_ids][:n_test], x[perm_ids][n_test:]
        y_test, y_train = y[perm_ids][:n_test], y[perm_ids][n_test:]
        z_test, z_train = z[perm_ids][:n_test], z[perm_ids][n_test:]
		
        # Train a decision tree to predict y using x and z.
        best_tree_x.train(concat(x_train, z_train), y_train)
    	mses_x.append(
            mse(best_tree_x.predict(concat(x_test, z_test)), y_test))
            
        # Train a decision tree to predict y using only z.
        best_tree_nox.train(z_train, y_train)&\label{lin:z2}&
        mses_nox.append(mse(best_tree_nox.predict(z_test), y_test))&\label{lin:z3}&
        
    # Obtain the one-tailed p-value for the null hypothesis that
    # on average, mses_nox is the same as mses_x.
    t, pval = ttest_1samp(mses_nox - mses_x)
    if t < 0:
    	return 1 - pval / 2.
    else:
    	return pval / 2.
\end{lstlisting}
\end{algorithm}

\section{Evaluating Statistical Test Performance}
An ideal nonparametric CIT should be fast, accurate, and generally applicable. To provide a thorough analysis of these factors we found it necessary to deviate from previous work in the evaluation method, as described in this section. We discuss alternative criteria and the problems we found with them in Sec.~\ref{sec:discussion_eval}. Our evaluation criteria are as follows:

\begin{compactenum}
\item[\textbf{Power}:] The test should have high power. Whenever $X\notindependent Y \mid Z$, the p-value should be small and should tend to 0 as sample sizes grows. 
\item[\textbf{Size}:] The size of the test should be equal to \emph{or lower than} its level $\alpha$. That is, whenever $X \independent Y \mid Z$, the p-value of the test should not be small for any sample size. Note that in some applications it might be important for the size of the test to be equal (neither lower nor higher) to its level. We discuss this further in Sec.~\ref{sec:discussion_eval}.
\item[\textbf{Speed}:] The test should be fast. This is motivated by the use cases discussed in Sec.~\ref{sec:motivation}.
\item[\textbf{Breadth of Application}:] The test should remain fast for large sample sizes and high-dimensional variables. It should have high power against any type of dependence, and low size for any distribution.
\end{compactenum}

\subsection{Evaluation method}
In order to understand the speed, accuracy and breadth of FIT and its competitors, we first chose four settings, described in Sec.~\ref{sec:datasets}. These four settings cover a range of possible data: low- and high-dimensional data, simple and complex functional relationships between the variables, continuous and mixed discrete-continuous systems\footnote{We did not consider systems where $X, Y, Z$ are all categorical as this case is easily solved without special conditional independence tests, as discussed in Sec.~\ref{sec:alternatives}} -- Table~\ref{tbl:datasets} summarizes the characteristics of each setting. We designed each setting in a way that enables us to vary its dimensionality or complexity. For convenience of plotting the results, we chose nine instantiations of the dimensionality/complexity parameters in each setting, and for each such instantiation, we generated a pair of datasets, one version where $X\independent Y \mid Z$ and one where $X\notindependent Y \mid Z$. 

We applied each candidate test described in Sec.~\ref{sec:alternatives} to each dataset, no matter the dimensionality or complexity. To further  understand the change in speed and performance of the tests as the number of samples grows, we ran each test on each dataset for \ttsamp{} (number of samples) varying between $10^2$ and $10^5$ logarithmically. For each \ttsamp{} we plot the $p$-value corresponding to the dependent and independent versions of the data, as well as the runtime of each test on each of the datasets and \ttsamp{}. This results in plots such as those shown in Fig.~\ref{fig:specresults_fit}-~\ref{fig:specresults_cci}. These figures provide a very detailed picture of the tests' performance, which we discuss in Sec.~\ref{sec:results}. In addition, Fig.~\ref{fig:minires} reduces the results to answer the question ``If I give each test at most 1 minute time, what is the best Type I and Type II error it can achieve on each dataset?''. This question follows naturally from our need to run large numbers of CITs for causal discovery (Sec.~\ref{sec:motivation}).

\subsection{Datasets}
\label{sec:datasets}
\begin{figure}
\centering
\includegraphics[width=.8\textwidth]{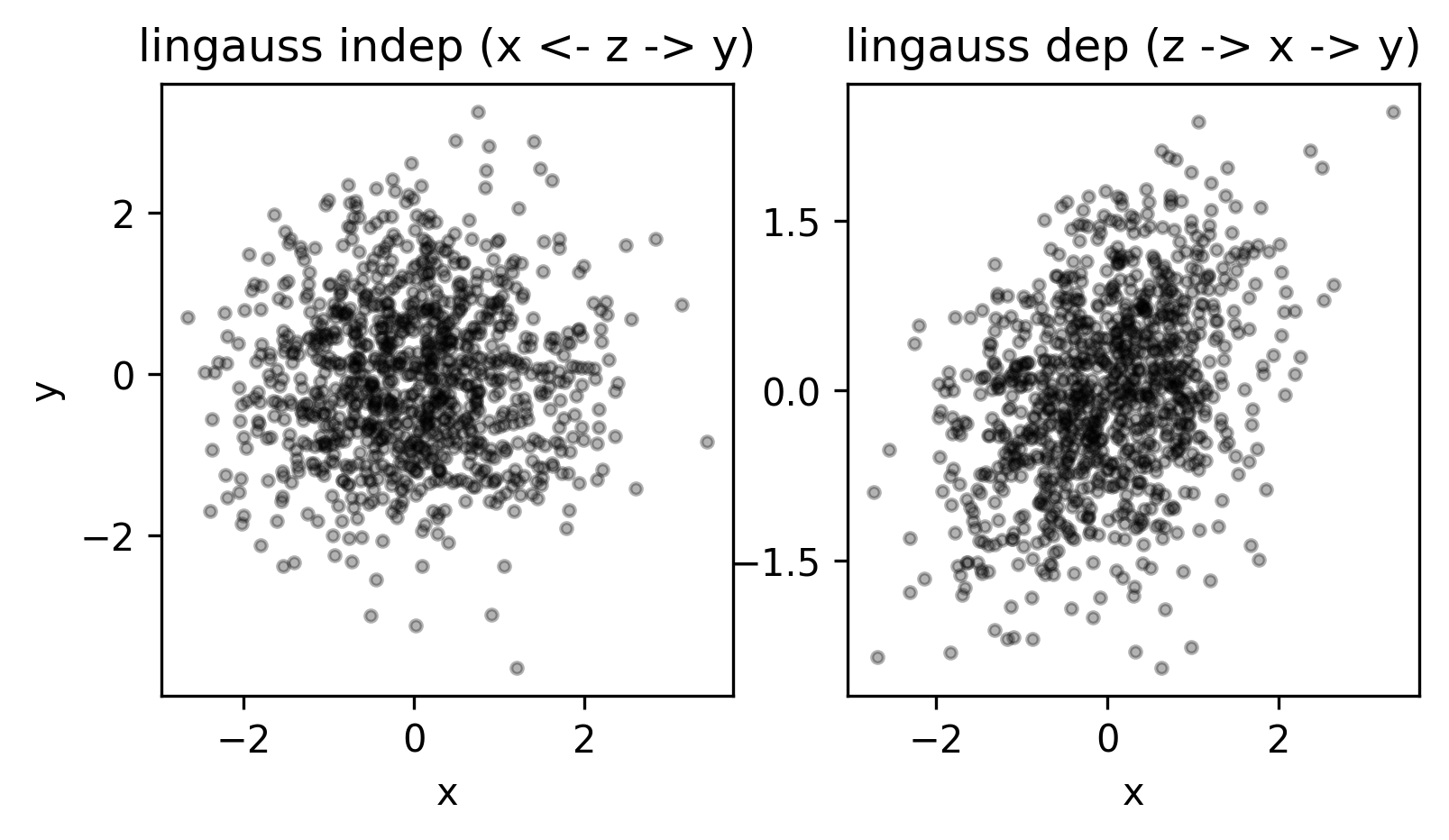}
\caption{An example \textsc{lingauss} dataset. For both the dependent and independent (1-dimensional) instantiation of the dataset we created $10^5$ random samples, and plotted $x$ vs $y$ that correspond to the cluster of $z$'s between the 50th and 51st percentile of $z$-values.}
\label{fig:dataset_lingauss}
\end{figure}

\begin{figure}
\centering
\includegraphics[width=.8\textwidth]{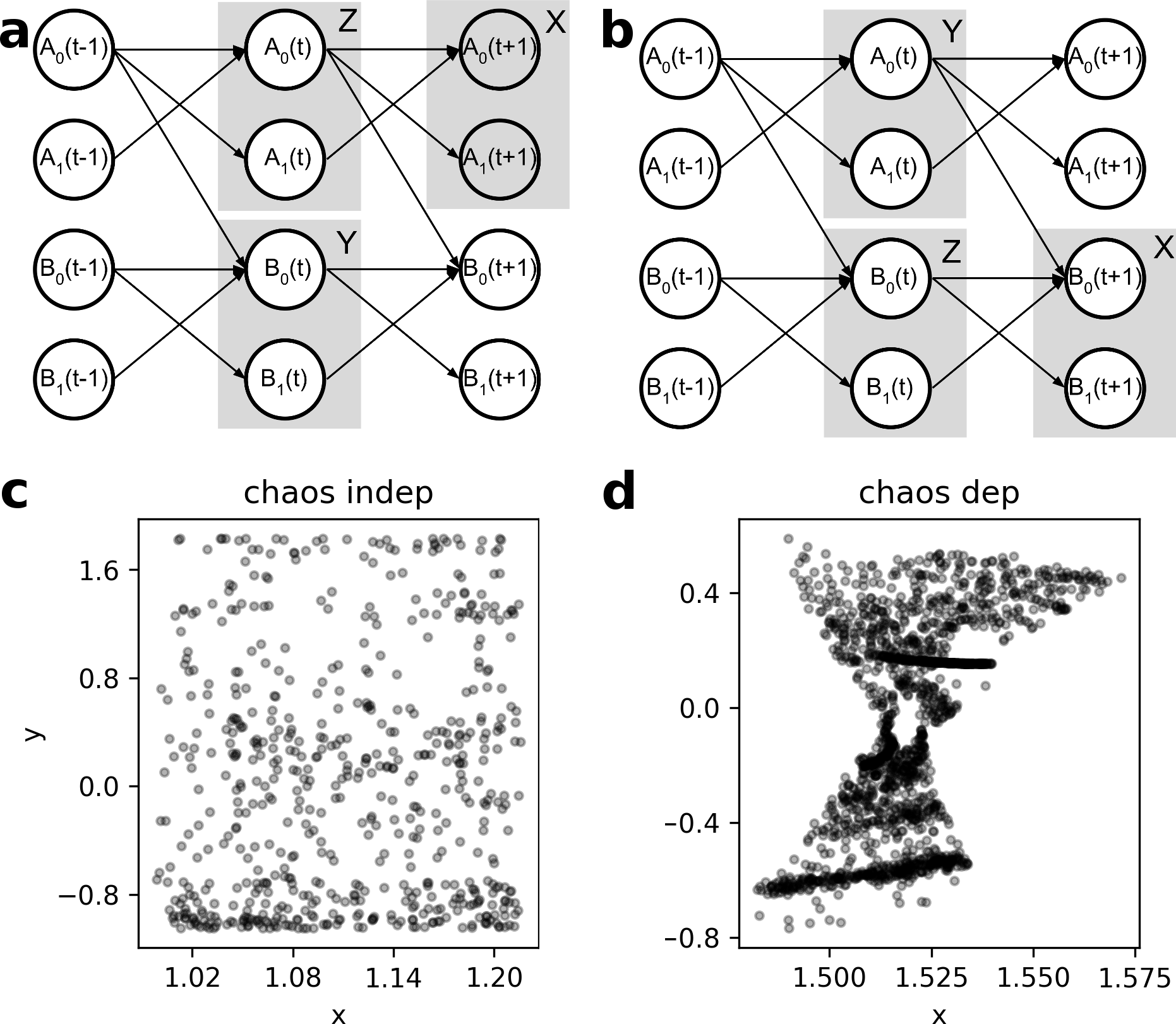}
\caption{Example \textsc{chaos} dataset when $\alpha = .5$. a) The generative model for the conditionally independent version of the dataset. Only three timesteps are shown and the third and fourth dimensions of $X$ and $Y$ are omitted, since they are independent noise variables. b) The generative model for the conditionally dependent version of the dataset. c) For the independent version of the dataset we created $10^5$ random samples. We then clustered the z's into clusters of size roughly 1000 using K-Means, and plotted $X_1$ vs.\ $Y_1$ corresponding to a randomly chosen cluster of z's (since $Z$ is multivariate continuous, the distribution $P(X, Y \mid Z \in \text{cluster of z values})$ is an approximation of $P(X, Y \mid z)$ for one value of $z$.) d) Same as c) but for the dependent data.}
\label{fig:dataset_chaos}
\end{figure}

\begin{figure}
\centering
\includegraphics[width=.8\textwidth]{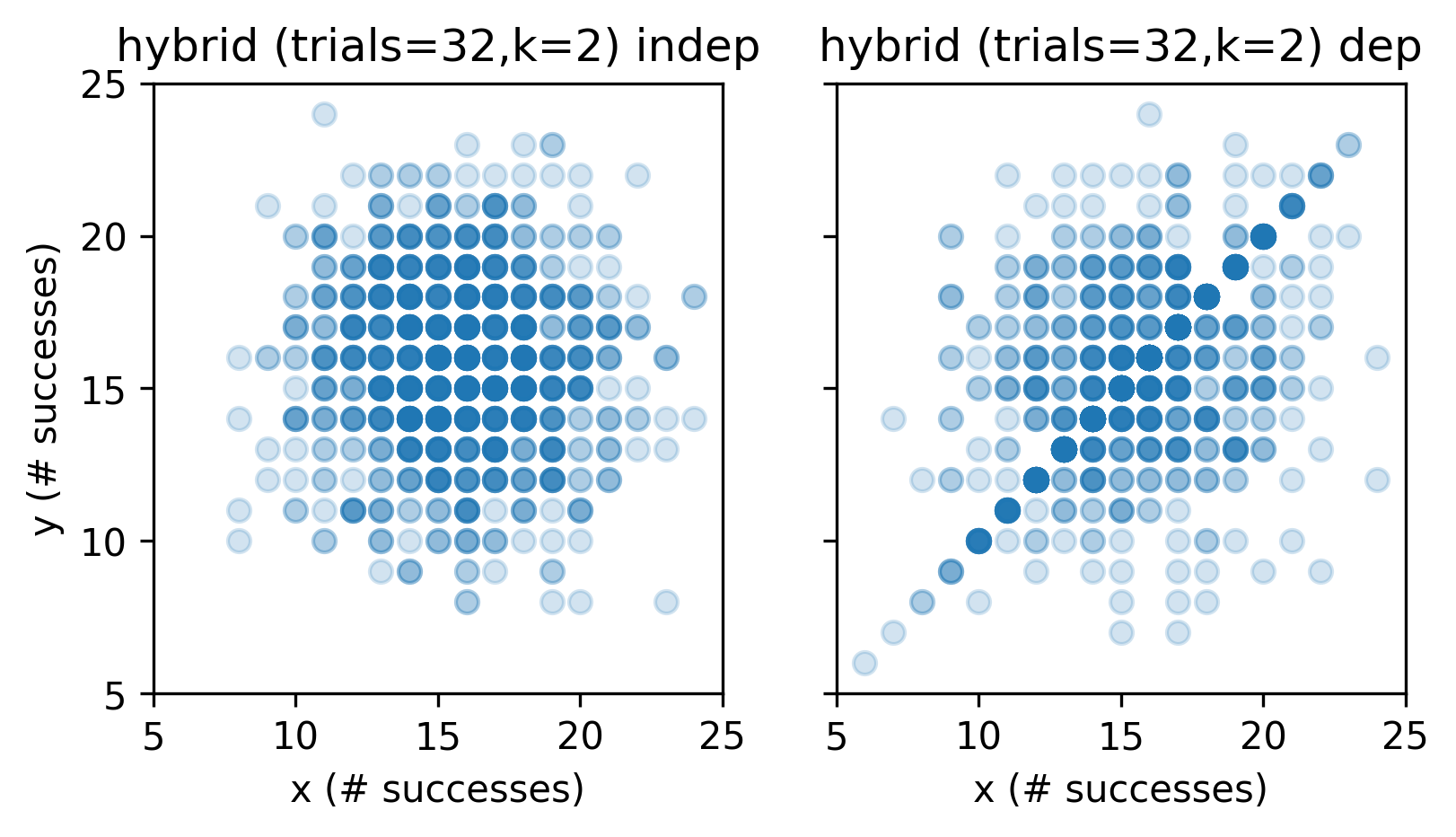}
\caption{Example \textsc{hybrid} dataset with $\gamma=2$ and \ttdim$=32$. For both the dependent and independent version of the dataset we created $10^5$ random samples, and picked the $x, y$ values that correspond to $z\in(.49, .51)$. The plot shows the heatmap (more saturated color = more samples) of the number of samples falling into given $x, y$ bin.}
\label{fig:dataset_hybrid}
\end{figure}

\begin{figure}
\centering
\includegraphics[width=.8\textwidth]{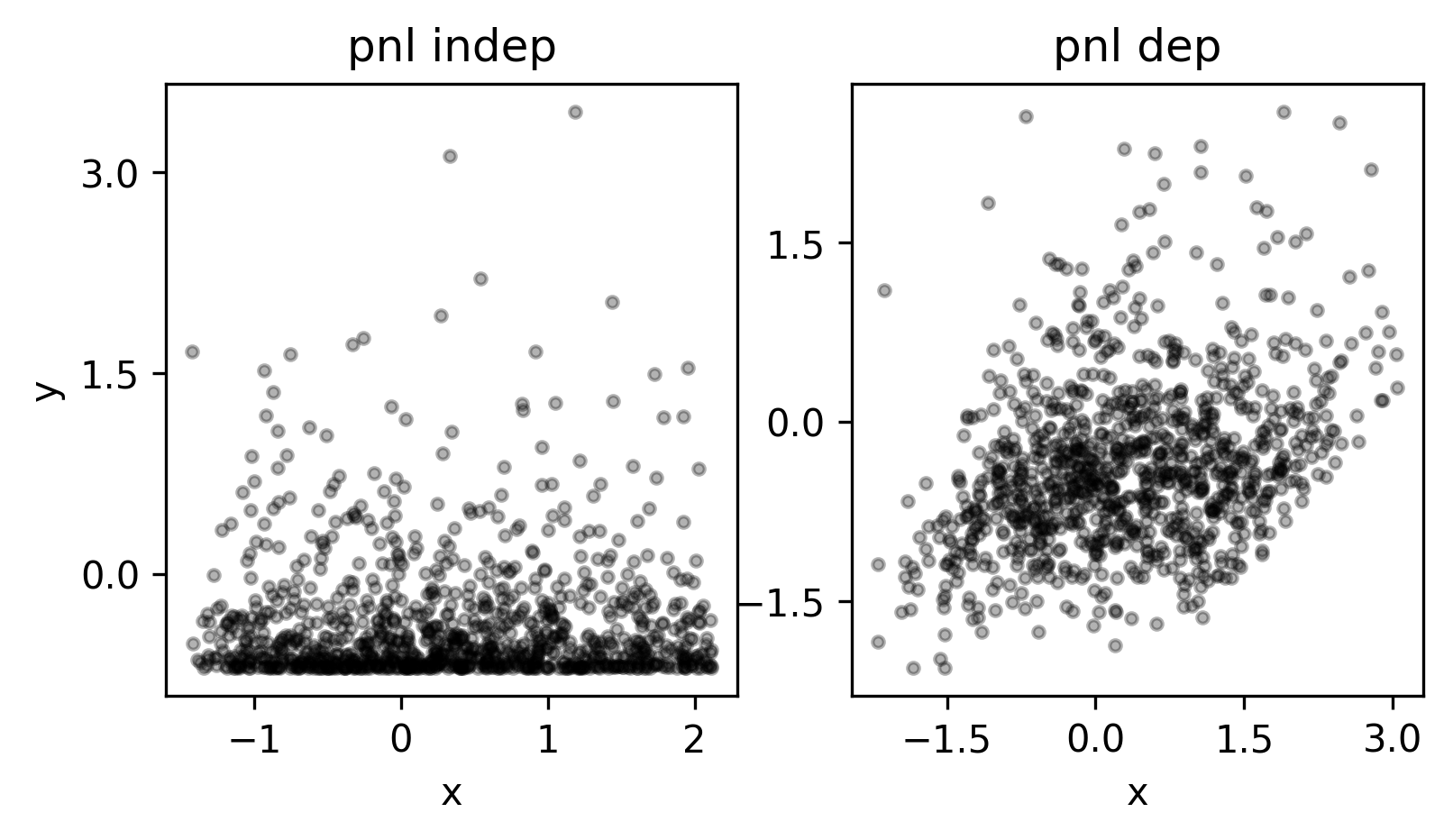}
\caption{Example \textsc{pnl} dataset. For both the dependent and independent (one-dimensional) version of the dataset we created $10^5$ random samples, and plotted $x$ vs.\ $y$ that correspond to the cluster of $z$'s between the 50th and 51st percentile of $z$-values.}
\label{fig:dataset_pnl}
\end{figure}

To evaluate FIT and compare it to alternatives, we used four settings to generate synthetic datasets, \textsc{LINGAUSS}, \textsc{CHAOS}, \textsc{HYBRID} and \textsc{PNL}, that cover a broad variety of use cases -- see Table~\ref{tbl:datasets}. We here provide descriptions of these settings in some detail to give a better sense of what the data can look like. Figures~\ref{fig:dataset_lingauss}-\ref{fig:dataset_pnl} show \emph{low dimensional examples} from these settings in order to provide at least some illustration of their difficulty and diversity.

\begin{table}
\centering
\begin{tabular}{lllll}\toprule
dataset&\textsc{lingauss}&\textsc{chaos}&\textsc{discrete}&\textsc{pnl}\\
\midrule
dimensionality&3-768&10&10-2080&3-258\\
type&x, y, z cont& x, y, z cont& x, y categorical; z cont&x, y, z cont\\
complexity&low&med-high&low-high&med\\
\bottomrule
\vspace{.05in}
\end{tabular}
\caption{Summary of evaluation settings. Dimensionality is summed over x, y and z. Type differentiates between continuous (cont) and categorical variables. Complexity is a non-rigorous quantification of the complexity of the conditional distributions that generate the data. See Sec.~\ref{sec:datasets} for details.}
\label{tbl:datasets}
\end{table}

\paragraph{\textsc{lingauss}:} Any reasonable statistical independence test should work on linear Gaussian models. Surprisingly, previous work was not evaluated on this basic case. For the $X\independent Y \mid Z$ version of the data, given \ttdim{} (the desired data dimensionality), we sample \ttsamp{} samples from the linear-Gaussian graphical model $X \leftarrow Z \rightarrow Y$:
\begin{compactenum}
\item Let $Z$ be a \ttdim-dimensional independent standard Gaussian, i.e. $Z \sim N(\mathbf{0}, \mathbf{I})$, where $\mathbf{I}$ is a $\ttdim{}\times\ttdim{}$ identity matrix.
\item Sample \ttdim$\times$\ttdim{} independent standard normal variables and arrange them into a \ttdim$\times$\ttdim{} matrix $A$ (for the $ZX$-"edge coefficient" matrix).
\item Sample \ttdim{}$\times$\ttdim{} independent standard normal variables and arrange them into a \ttdim$\times$\ttdim{} matrix $B$ (for the $ZY$-"edge coefficient" matrix).
\item Let $X = AZ + $\ttdim-dimensional independent standard Gaussian.
\item Let $Y = BZ + $\ttdim-dimensional independent standard Gaussian.
\end{compactenum}

For the $X\notindependent Y \mid Z$ case, given \ttdim{} (the desired data dimensionality), we sample \ttsamp{} samples from the linear-Gaussian graphical model $Z \rightarrow X \rightarrow Y$:
\begin{compactenum}
\item Let $Z$ again be a \ttdim-dimensional independent standard Gaussian.
\item Sample \ttdim$\times$\ttdim{} independent standard normal variables and arrange them into a \ttdim$\times$\ttdim{} matrix $A$ (for the $ZX$-"edge coefficient" matrix).
\item Sample \ttdim{}$\times$\ttdim{} independent standard normal variables and arrange them into a \ttdim$\times$\ttdim{} matrix $B$ (for the $XY$-"edge coefficient" matrix).
\item Let $X = AZ + $\ttdim-dimensional independent standard Gaussian.
\item Let $Y = BX + $\ttdim-dimensional independent standard Gaussian.
\end{compactenum}
Note that we deliberately did \emph{not} create the (conditionally) dependent dataset by adding an extra "edge" between $X$ and $Y$ to the conditionally independent "common cause model" in order to ensure that the dimensionalities of the relation between $X$ and $Y$ conditional on $Z$ remain the same between the independent and dependent case.

We evaluated the CITs using \textsc{LINGAUSS} with \texttt{dim} = 1, 2, 4,... , 256.

\paragraph{\textsc{chaos}:} \citet{doran_permutation-based_2014} used this dataset to evaluate their CIT. The data is sampled from a highly nonlinear, chaotic dynamical system (see Fig.~\ref{fig:dataset_chaos}). The dimensionality of the data is fixed: $X$ and $Y$ both have \ttdim{} = 4 and $Z$ has \ttdim{} = 2. A scalar parameter $\alpha\in (0, 1)$ controls the complexity of the dataset. Let $A\in\mathbb{R}^2, B\in\mathbb{R}^2$ follow the equations (through time index $t$):
\begin{align*}
A(t, 0) &= 1.4 - A(t-1, 0)^2 + .3 A(t-1, 1)\\
A(t, 1) &= A(t-1, 0)\\
B(t, 0) &= 1.4 - (\alpha A(t-1, 0) B(t-1, 0) + (1-\alpha) B(t-1, 0)^2) + .1 B(t-1, 1)\\
B(t, 1) &= B(t-1, 0)
\end{align*}
To create the $X\independent Y \mid Z$ data, take \begin{align*}
X(t) &= A(t+1),\\
Y(t) &= B(t),\\
Z(t) &= A(t). 
\end{align*}
To create the $X\notindependent Y \mid Z$ data, take 
\begin{align*}
X(t) &= B(t+1),\\
Y(t) &= A(t),\\
Z(t) &= B(t).
\end{align*} 
Finally, to make the task more difficult, append two independent Gaussian variables with $N(0, 0.5)$ as the third and fourth dimension of $X$ and $Y$. To make the samples independent, we generated $10^6$ samples and chose a random subset of the desired size to form our datasets.
Fig.~\ref{fig:dataset_chaos}a,b illustrate the choices of $X,Y,Z$ in the generative model. 

We evaluated the CITs using \textsc{CHAOS} with $\alpha$ = .01, .04, .16, .32, .5, .68, .84, .96, .99.

\paragraph{\textsc{hybrid}:} We created this hybrid continuous-categorical dataset as we have not seen any previous methods evaluated on hybrid data (see Sec.~\ref{sec:motivation}). We deliberately constructed \emph{categorical} variables, as opposed to (e.g.\ numerical) discrete variables, whose values still have some order.

Given a count parameter $\gamma$, each sample for a \textsc{hybrid}($\gamma$)-dataset, say $(x_i, y_i, z_i)$, was constructed using the following pseudocode:

\begin{compactenum}
\item Let $z_i$ be a a \ttdim-dimensional continuous vector sampled from the uniform Dirichlet distribution with $Z\sim \text{Dirichlet}(\mathtt{dim})$.
\item Draw two independent samples $S_X$ and $S_Y$ from the multinomial distribution with probability parameter $z_i$ and count parameter $\gamma\in\mathbb{N}$. 
\item Since we want $x_i$ and $y_i$ to represent multi-dimensional categorical variables, convert each dimension of $S_X$ and $S_Y$ to the one-hot encoding. For example, take \ttdim=3 and $\gamma = 2$. Suppose $z_i = [.1, .6, .3]$ for a particular sample. Then $S_X$ might be $(0,2,0)$, corresponding to zero draws in the first and third bucket and two draws in the second bucket of the sample from the $z_i$-multinomial. Then the one-hot encoding of $S_X$ makes $x_i = [0, 0, 0, 0, 0, 1, 0, 0, 0]$. 
The one-hot encoding enforces the discrete metric on the $X$-space.
\item (If creating the dependent version of the dataset), toss an unbiased coin. If heads, then overwrite the value of $y_i$ with the value of $x_i$.
\end{compactenum}

Since $Z$ is a sufficient statistic of the distributions of $X$ and $Y$, we have $X \independent Y \mid Z$ in Steps 1-3, and Step 4 is used to create an obvious dependency (which nevertheless turns out to be non-trivial to detect by many of the algorithms, see Sec.~\ref{sec:results}). 

We evaluated the CITs using \textsc{HYBRID} with $(\gamma, \ttdim)$ = (2, 2), (2, 8), (2, 32), (8, 2), (8, 8), (8, 32), (32, 2), (32, 8), (32, 32).

\paragraph{\textsc{pnl}:} Various versions of this dataset have been used before in~\citep{zhang_kernel-based_2012,doran_permutation-based_2014,strobl_approximate_2017}. We used the version from the most recent article~\citep{strobl_approximate_2017}. Since this version offers only scalar $X, Y, Z$, we extended it to the multi-dimensional case in a straightforward way. To create the generating mode for the \ttdim-dimensional \textsc{PNL} dataset:
\begin{compactenum}
\item Let $A$ be a \ttdim$\times$\ttdim-dimensional matrix of independent standard normal scalars.
\item Let $Z$ be a \ttdim{}-dimensional Gaussian with a covariance $A\times A^T$ and mean 0.
\item Let $X$ be a random function of the first coordinate of $Z$ (see below).
\item Let $Y$ be a (re-sampled) random function of the first coordinate of $Z$ (see below).
\item (To create the dependent version of the dataset,) add a Gaussian variable with mean zero and standard deviation .5 to both $X$ and $Y$.
\end{compactenum}

This procedure uses ``random functions''. To obtain such functions, following~\citet{strobl_approximate_2017}, we sample with equal probability from the set of functions $\{v\mapsto v, v\mapsto v^2, v\mapsto v^3, v\mapsto \tanh{v}, v\mapsto \exp{(-|v|)}\}$.

Note that in this dataset, $X$ and $Y$ are always one-dimensional. We evaluated the CITs using \textsc{PNL} with \ttdim = 1, 2, 4, ..., 256.

\section{Empirical Results and Comparison with Previous Work}
\label{sec:results}
\begin{figure}[h!]
\centering
\includegraphics[width=1.\textwidth]{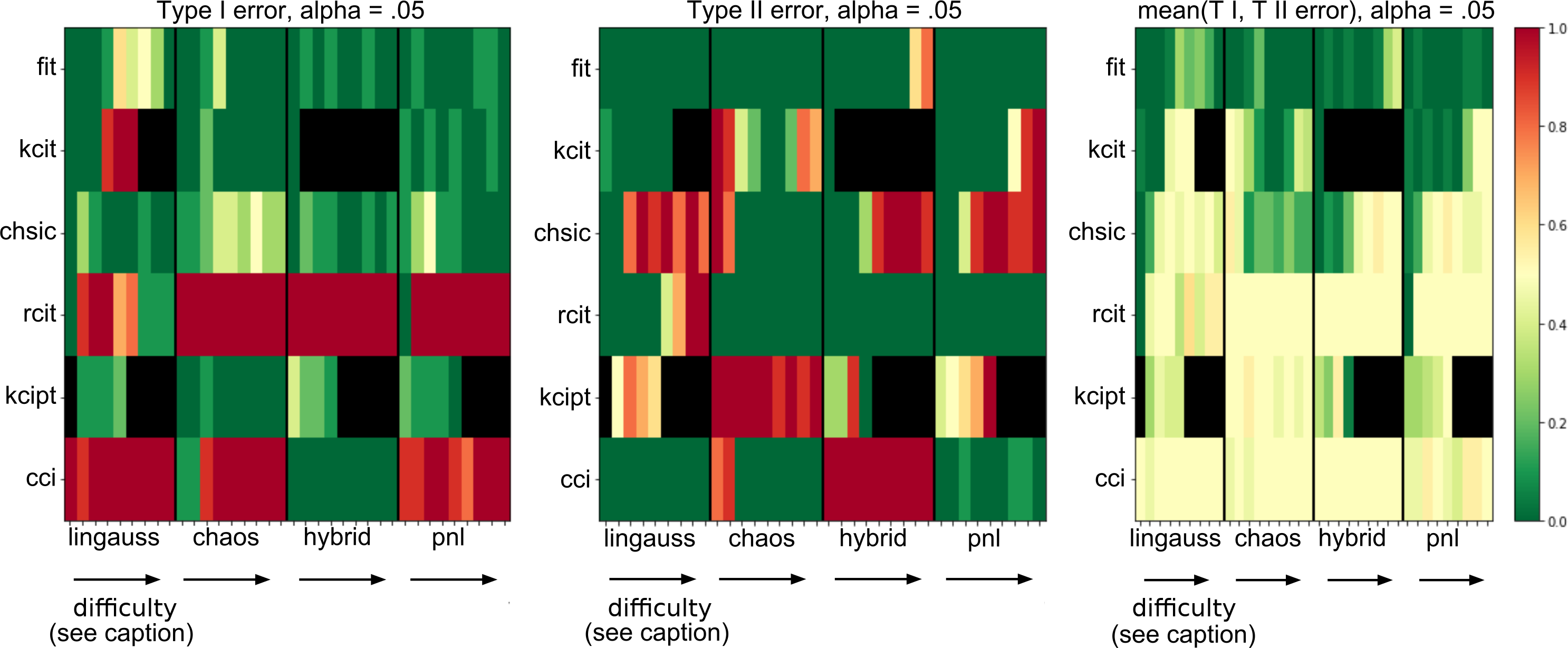}
\caption{Type I and II errors at significance level $\alpha=.05$. For each dataset setting (x-axis) and each dataset difficulty (narrow columns within each x-axis block), we computed the errors each method (y-axis) returns on the largest number of samples it can process within 60s (colors indicate error magnitude). Black regions correspond to the cases where an algorithm was not able to process even 100 samples within 100s. Black vertical lines separate the four settings, and for each setting the columns are arranged in increasing difficulty, e.g. for \textsc{lingauss} the first column corresponds to \ttdim=1, last column to \ttdim=256. See Figures~\ref{fig:result_chain_fit}-\ref{fig:result_pnl_cci} for full results plots, covering all sample sizes and all $p$-values.}
\label{fig:minires}
\end{figure}

For each out of the 72 dataset versions (4 settings times 9 parameter instantiations for both the dependent and independent version), we obtain $p$-values of FIT and each of the five conditional independence tests described in Sec.~\ref{sec:alternatives}. In each case, we varied the number of samples between $10^2$ and $10^5$ logarithmically. In addition, we stopped any test that exceeded the time threshold of about 100s. In this section, for each CIT we picked the results plot that best summarizes the broad behavior of the test across the datasets. In addition, Figure~\ref{fig:minires} provides a summary view of the full results. Due to their sheer volume we leave the detailed full results to Appendix~\ref{sec:full_res}.   Nevertheless, the full results plots form an important part of the evidence that motivates the discussion in this section.

\FloatBarrier
For all methods, as expected, lower dataset difficulty results in smaller Type I and higher Type II errors. The average of the two error types, shown in the right column of Figure~\ref{fig:minires}, shows that FIT achieves the best performance across the board. The remaining methods either have high errors (CHSIC, RCIT, CCI) or are too slow to process a large proportion of the datasets (KCIT, KCIPT). More detailed remarks on each method's performance follow.

\subsection{FIT}

\begin{figure}[h!]
\centering
\includegraphics[width=1.\textwidth]{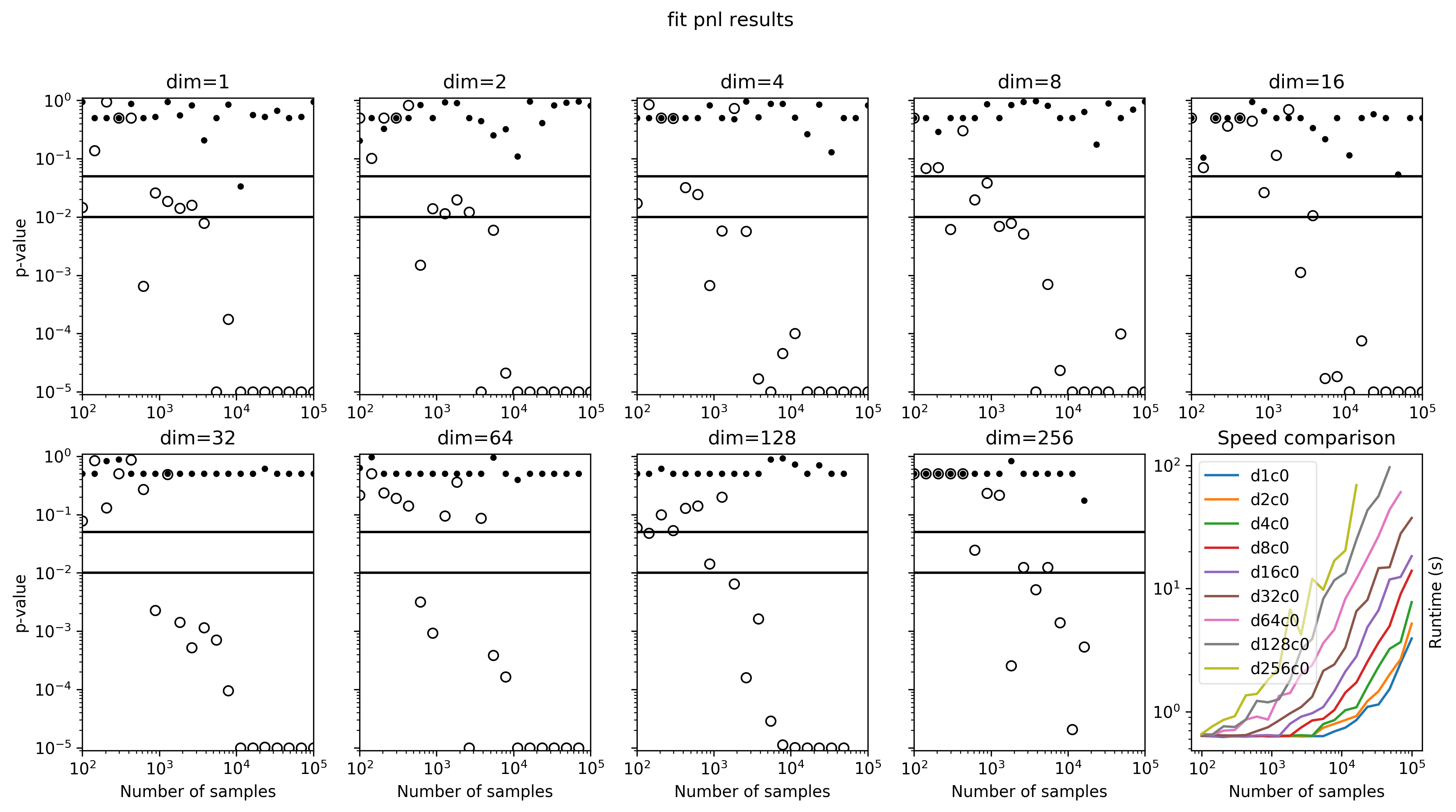}
\caption{Typical FIT results: all the $p$-values plotted against sample size on log-log scales. Here we show as example the results on \textsc{PNL} data, one plot for each dimensionality setting. Black dots correspond to dataset versions where $X\independent Y \mid Z$. To achieve low Type I error, they should be close to 1. White dots correspond to dataset versions where $X \notindependent Y \mid Z$. To achieve low Type II error, they should be close to 0. (Circled black dots result when the independent and dependent version of the dataset returned the same $p$-value.) The values are clipped at $10^{-5}$. The horizontal lines mark the $p$-values of $0.05$ and $0.01$, respectively. Lack of datapoints indicates that the method ran out of time (e.g.\ for high sample sizes in the highest dimensionality datasets). The last plot shows the runtime for different sample sizes on each of the datasets. We stopped the methods after 100s.}
\label{fig:specresults_fit}
\end{figure}

Figure~\ref{fig:specresults_fit} illustrates typical behavior of the test, in this case applied to the \textsc{PNL} dataset. The only data that causes problems for the FIT algorithm is the high-dimensional, sparse versions of the \textsc{hybrid} dataset. None of the other methods succeed here either (see full result plots in Appendix~\ref{sec:full_res}). The results suggest the following considerations for the FIT test:
\begin{compactenum}
\item The test can fail for small sample sizes. It becomes reliable once the sample size reaches $10^3-10^4$.
\item The test is fast even for a large number of dimensions and sample sizes, most often returning in less than 10s.
\item Type I and Type II errors decrease consistently as the number of samples increases.
\end{compactenum}

\FloatBarrier
\newpage
\subsection{CHSIC}
\begin{figure}[h!]
\centering
\includegraphics[width=1.\textwidth]{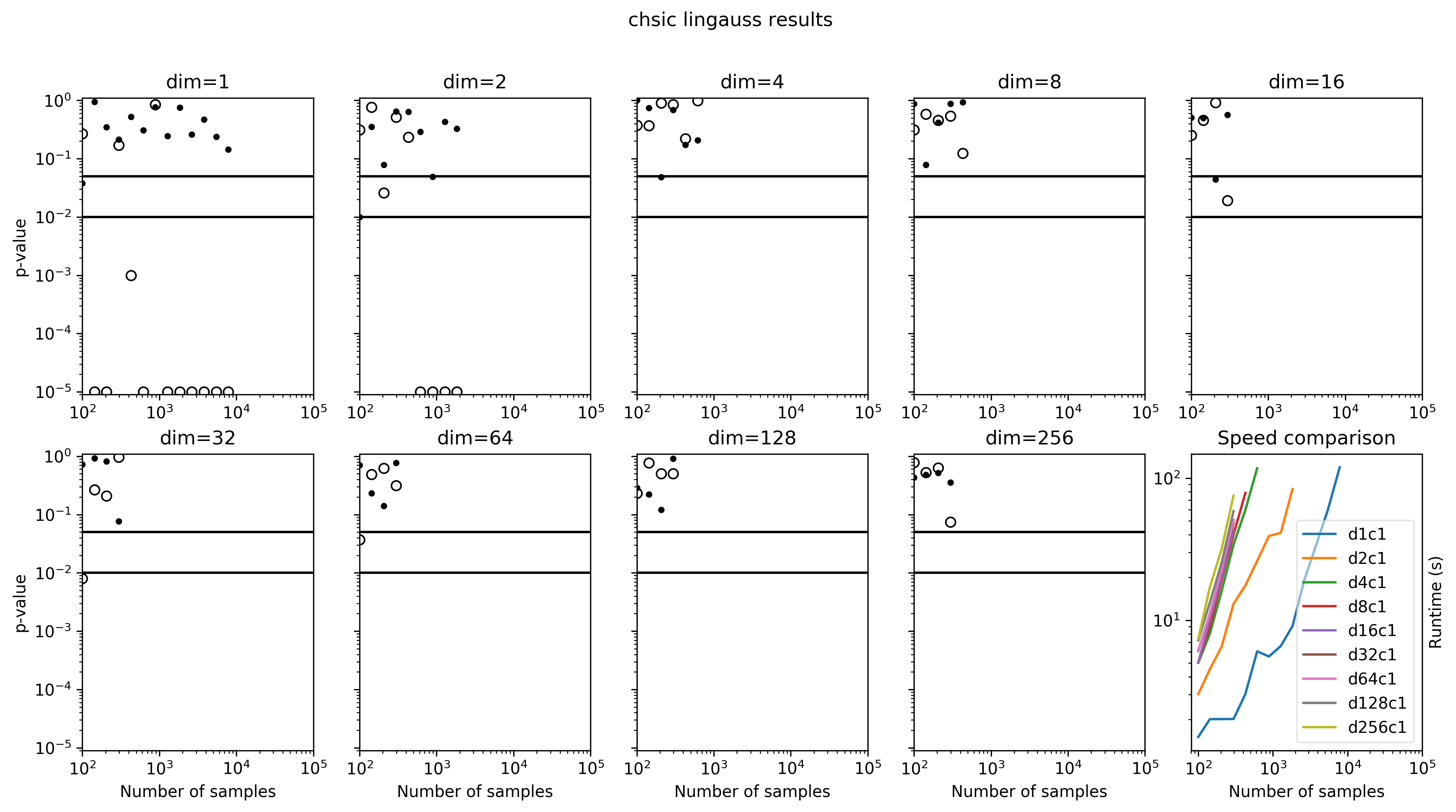}
\caption{Typical CHSIC results (shown here on the \textsc{lingauss} datasets). Black dots correspond to dataset versions where $X\independent Y \mid Z$. To achieve low Type I error, they should be close to 1. White dots correspond to dataset versions where $X \notindependent Y \mid Z$. To achieve low Type II error, they should be close to 0. The values are clipped at $10^{-5}$. Lack of datapoints indicates that the method ran out of time. Note the comparison to the runtime plot in Fig.~\ref{fig:specresults_fit}.}
\label{fig:specresults_chsic}
\end{figure}

CHSIC performs well on low-dimensional, low-sample-size and low-difficulty data, but has trouble detecting conditional dependence in the more difficult cases. Fig.~\ref{fig:specresults_chsic} illustrates this using the \textsc{lingauss} datasets. Overall, our results suggests that CHSIC:
\begin{compactenum}
\item Achieves very good Type I and Type II error levels for most datasets settings, but only for the lowest difficulty settings.
\item Fails to return meaningful results for sample sizes over 1000 and/or large difficulty data.
\end{compactenum}

\FloatBarrier
\subsection{KCIT}
\begin{figure}[h!]
\centering
\includegraphics[width=1.\textwidth]{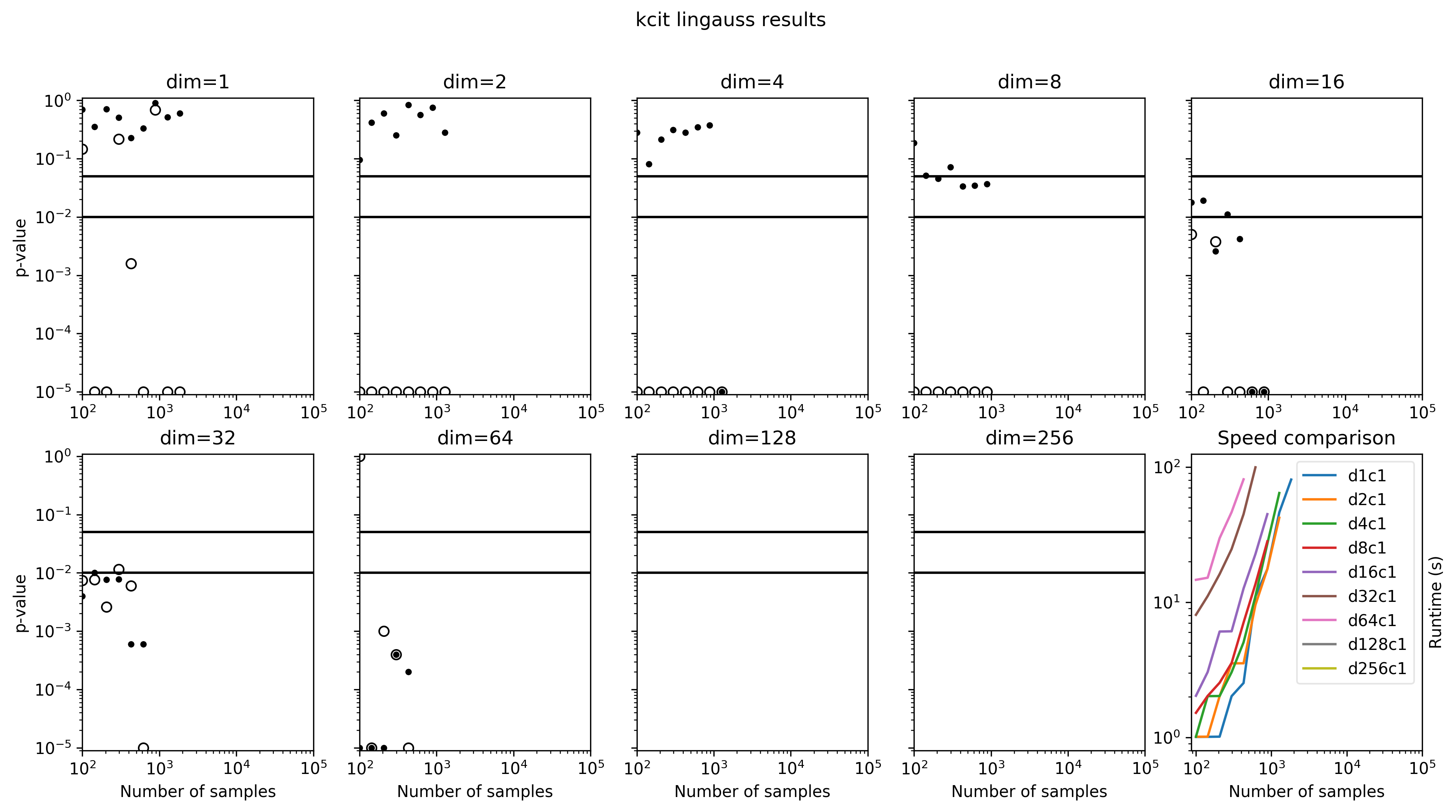}
\caption{Typical KCIT results (shown here on the \textsc{lingauss} datasets). Black dots correspond to dataset versions where $X\independent Y \mid Z$. To achieve low Type I error, they should be close to 1. White dots correspond to dataset versions where $X \notindependent Y \mid Z$. To achieve low Type II error, they should be close to 0. The values are clipped at $10^{-5}$. Lack of datapoints indicates that the method ran out of time -- almost always when number of samples was over 1000.}
\label{fig:specresults_kcit}
\end{figure}

KCIT does well in most cases when it returns, but is often too slow to approach the harder datasets -- as illustrated in Fig.~\ref{fig:specresults_kcit}, again using the \textsc{lingauss} datasets.
\begin{compactenum}
\item KCIT achieves great Type I and Type II errors for low dimensionality / low difficulty datasets.
\item Type I error of the method increases as dataset difficulty increases, while Type II error stays low, across all the settings.
\item The method's speed only allows it to tackle low-dimensionality data with sample size smaller than 1000.
\end{compactenum}

\FloatBarrier
\subsection{RCIT}
\begin{figure}[h!]
\centering
\includegraphics[width=1.\textwidth]{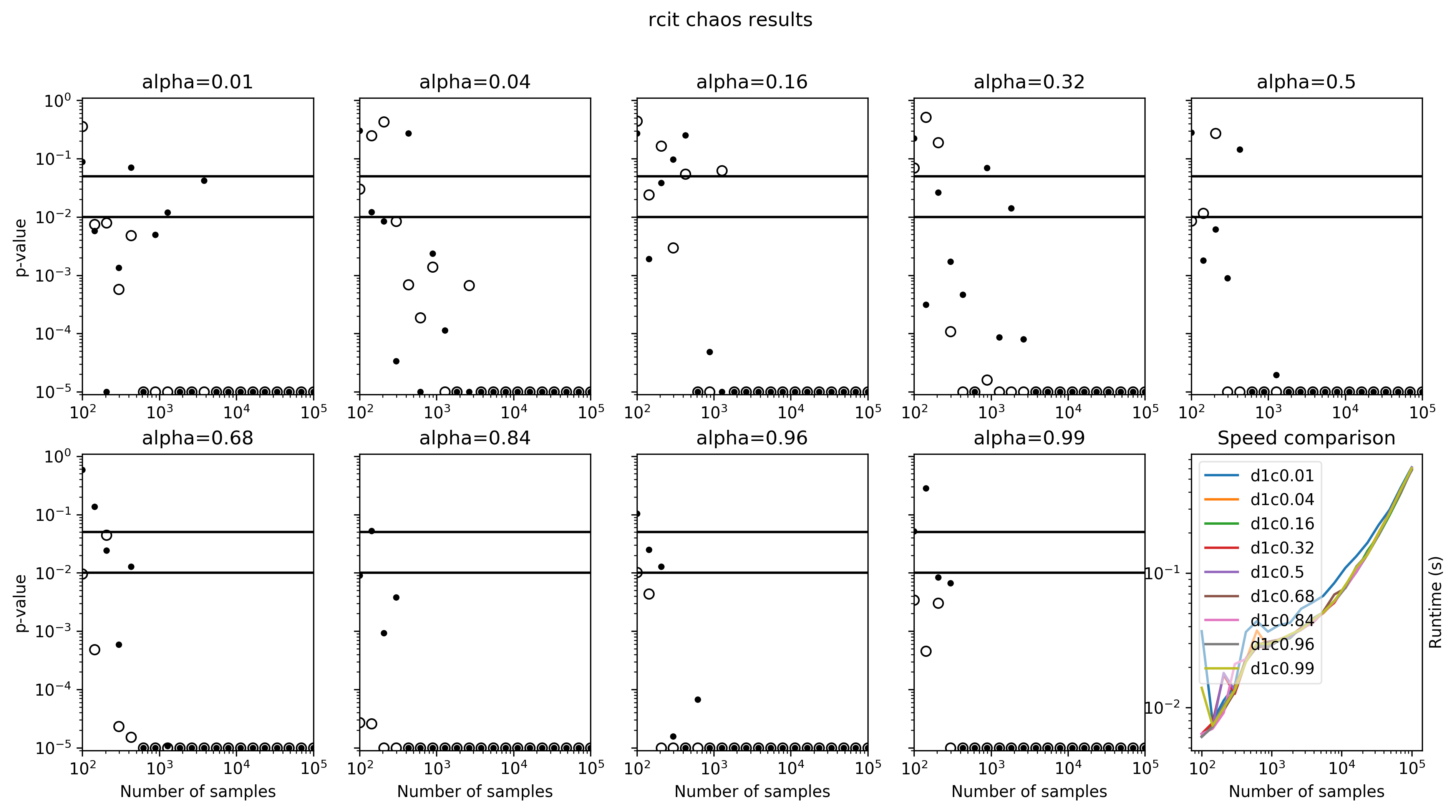}
\caption{Typical RCIT results (shown here on the \textsc{chaos} datasets). Black dots correspond to dataset versions where $X\independent Y \mid Z$. To achieve low Type I error, they should be close to 1. White dots correspond to dataset versions where $X \notindependent Y \mid Z$. To achieve low Type II error, they should be close to 0. The values are clipped at $10^{-5}$. Lack of datapoints indicates that the method ran out of time.}
\label{fig:specresults_rcit}
\end{figure}

In almost all cases (with the only exception of 1-dimensional \textsc{lingauss} and 1-dimensional \textsc{pnl} data), RCIT returns very low $p$-values as the number of samples increases, thus yielding high Type I errors. Fig.~\ref{fig:specresults_rcit} illustrates this behavior.

\FloatBarrier
\subsection{KCIPT}
\begin{figure}[h!]
\centering
\includegraphics[width=1.\textwidth]{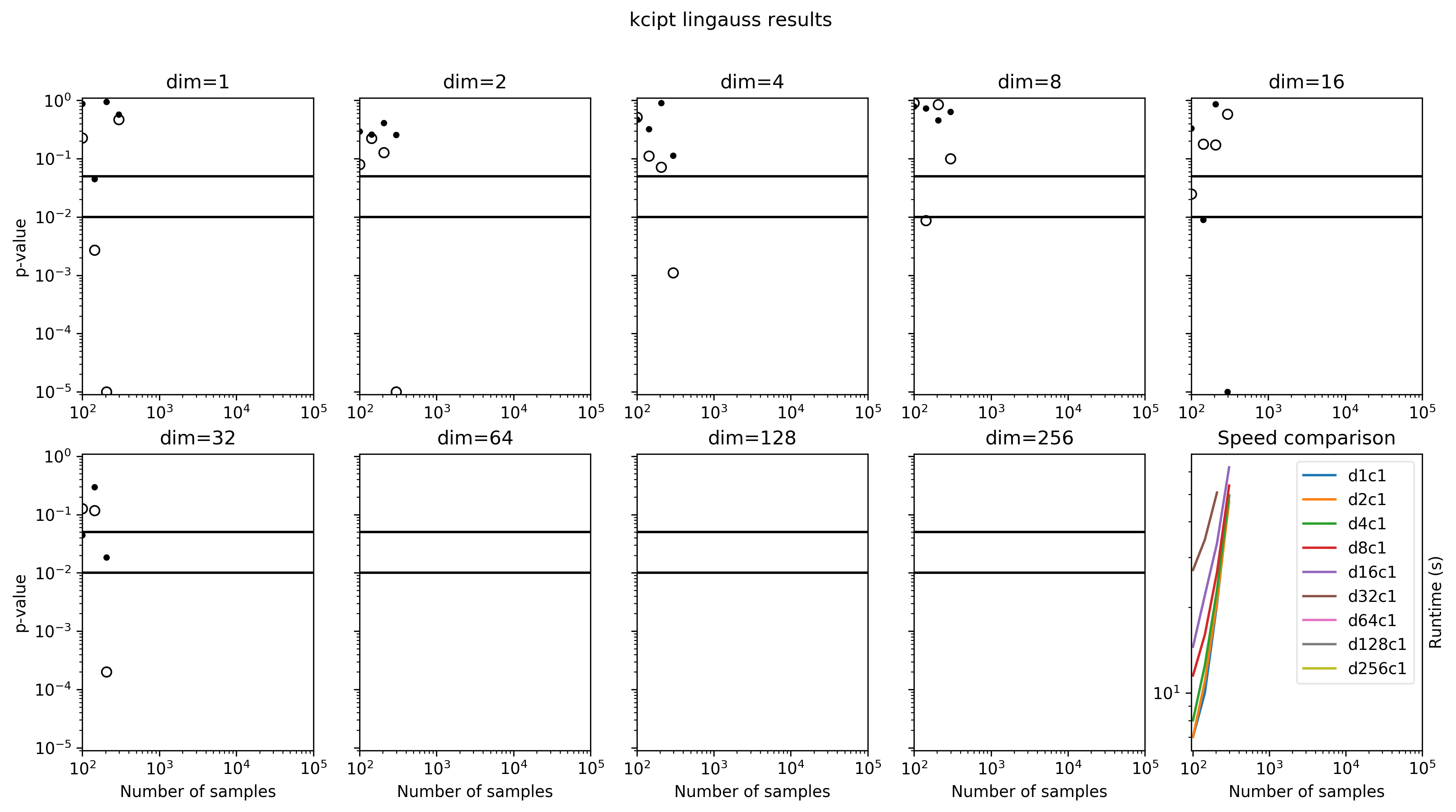}
\caption{Typical KCIPT results (shown here on \textsc{lingauss} data). Black dots correspond to dataset versions where $X\independent Y \mid Z$. To achieve low Type I error, they should be close to 1. White dots correspond to dataset versions where $X \notindependent Y \mid Z$. To achieve low Type II error, they should be close to 0. The values are clipped at $10^{-5}$. Lack of datapoints indicates that the method ran out of time.}
\label{fig:specresults_kcipt}
\end{figure}

KCIPT fails on a majority of the datasets, and is too slow to process any but the least-dimensional and lowest-sample cases -- as shown in Fig.~\ref{fig:specresults_kcipt}.

\FloatBarrier
\subsection{CCI}
\begin{figure}[h!]
\centering
\includegraphics[width=1.\textwidth]{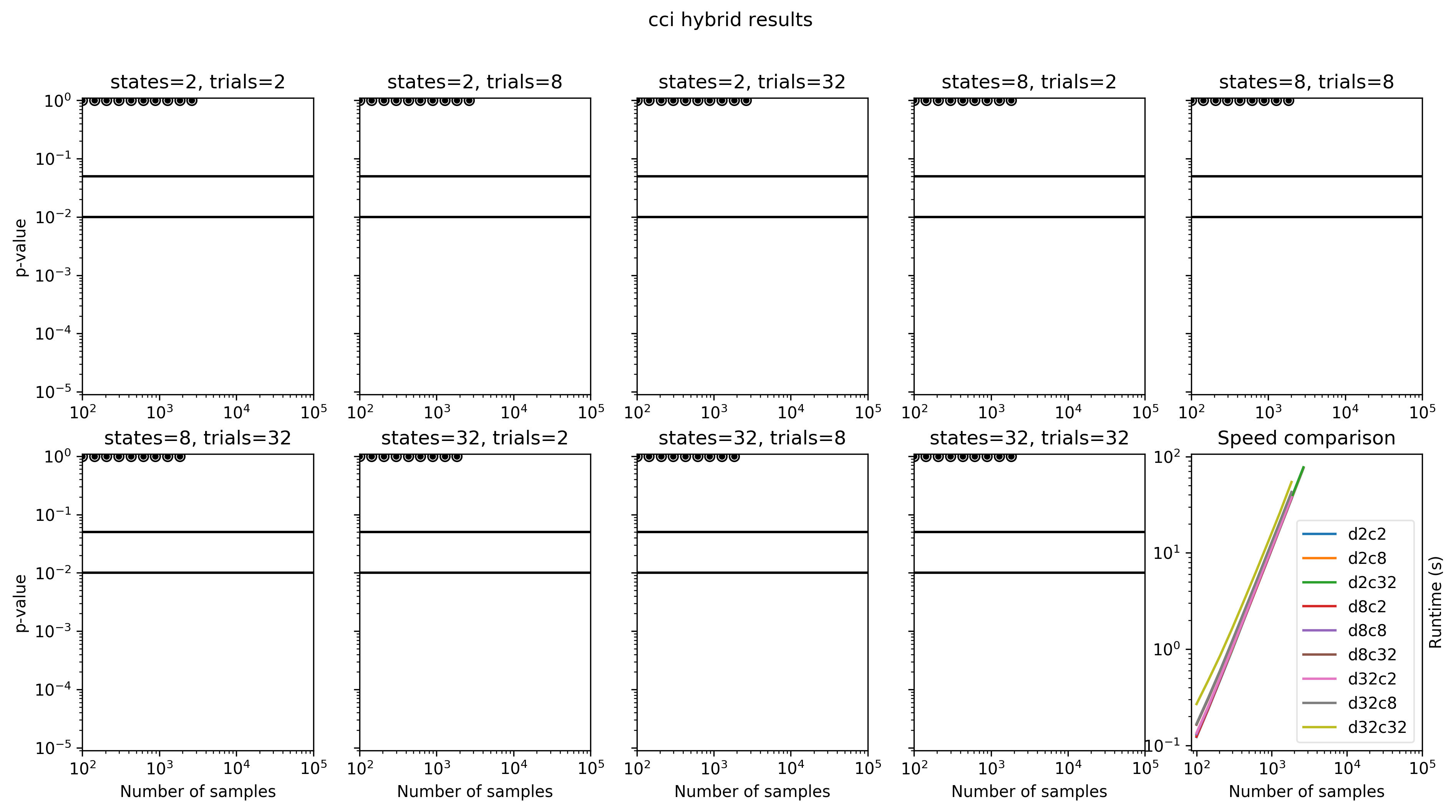}
%
\caption{Typical CCI results (shown here on \textsc{hybrid} data). Black dots correspond to dataset versions where $X\independent Y \mid Z$. To achieve low Type I error, they should be close to 1. White dots correspond to dataset versions where $X \notindependent Y \mid Z$. To achieve low Type II error, they should be close to 0. The values are clipped at $10^{-5}$. Lack of datapoints indicates that the method ran out of time. CCI returns only a binary result of 0 (for dependence) or 1 (for independence).}
\label{fig:specresults_cci}
\end{figure}

CCI fails on most datasets. In addition, instead of outputting a $p$-value it outputs a binary decision: either 1 (indicating the data is conditionally independent) or 0 (for conditionally dependent data). Unfortunately, in our evaluation it has a tendency to output 0 for most of the continuous datasets, and 1 for the \textsc{hybrid} dataset (as shown in Fig.~\ref{fig:specresults_cci}).

\section{Discussion}
\label{sec:discussion}
We showed that FIT is a fast and accurate conditional independence test. Our evaluation covered a broad range of dataset sizes, dimensionalities and complexities. In this evaluation, FIT performs as well as or better than the alternatives. In addition, it is the only algorithm that applies successfully to large sample sizes. In this section, we discuss alternatives to our evaluation strategy, and clarify some aspects of FIT.

\subsection{Alternative evaluation metrics}
\label{sec:discussion_eval}
Previous work used a variety of methods to evaluate the performance of CITs. \citet{fukumizu2008} and \citet{zhang_kernel-based_2012} report Type I and Type II errors on a number of datasets. This is similar to our condensed presentation in Fig.~\ref{fig:minires}. However, their evaluation considered a fixed, small sample size of 200 and 400 datapoints (as the evaluated methods are too slow to attack a larger dataset). In contrast, we allow each algorithm to process as many datapoints as possible with a fixed time limit.

As~\citet{doran_permutation-based_2014} point out however, Type I and Type II errors can bias the results. Some algorithms might perform particularly well at one significance level, and some applications might require non-standard significance levels. Thus,~\citet{doran_permutation-based_2014} proposed to condense the performance of a CIT across all significance levels to two statistics. First, their ``area under the power curve'' (AUPC) is the area under the cumulative distribution function of the $p$-values the algorithm returns on a conditionally dependent dataset. Second, they perform a test on the distribution of $p$-values: their ``KS-test $p$-value'' is the $p$-value of the Kolmogorov-Smirnoff test~\citep{stephens_edf_1974} for the null hypothesis that the $p$-values of a CIT are uniformly distributed on a conditionally independent dataset. A good test has low AUPC and a non-low KS-test $p$-value. 

The problem with both of these approaches is that they do not consider how the behavior of a CIT changes as the number of samples grows. To us, an important requirement is that for any type of data, the test is ``correct'' in the limit of an infinite number of samples. Looking at a limited range of sample sizes can be deceptive. Consider Fig.~\ref{fig:result_pnl_rcit} with dim=128. RCIT seems to be doing very well for sample sizes between 100 and 1000. However, it fails for any larger number of samples. In reality, both for the dependent and independent version of the dataset RCIT's $p$-values follow a decaying trend as the number of samples grows, which is not a correct behavior. 

Evaluating a CIT on a fixed sample size is deceptive in a similar way to evaluating it at a fixed significance level. The best (although energy-consuming) way we found to understand the behavior of a CIT and compare it with other algorithms is to carefully examine how its $p$-value changes with a changing sample size on a variety of datasets. Figures~\ref{fig:result_chain_fit}-~\ref{fig:result_pnl_cci} show these $p$-value plots. We encourage any reader interested in a thorough comparison of the CIT methods to take the time to examine these plots.

\subsection{Why use a Decision Tree Regression?}
\label{sec:discussion_ml}
Algorithm~\ref{alg:fit} uses decision tree regression (DTR) as its machine learning component. Other regression algorithms could serve as a substitute for DTR. There are two crucial requirements for such algorithms (apart from good predictive power): 1) computational efficiency and 2) ability to handle structured multi-dimensional inputs. Initially, we implemented FIT using a neural network as the back-end, efficiently implemented in Tensorflow~\citep{abadi_tensorflow:_2016}, and trained on a GPU (Titan Xp). However, given that we want FIT to apply to both small and large, high- and low-dimensional data, the hyperparameter choice for the neural network regressor becomes a complex problem -- for example~\citet{jaderberg_population_2017} propose an simple and effective hyperparameter search method that nevertheless assumes having access to a cluster of a hundred GPU machines as a reasonable setting. DTRs are fast and require us to choose the value of only one hyperparameter.

\subsection{Failure cases}
\label{sec:discussion_failure}
Our Fast (Conditional) Independence Test assumes that if $X\notindependent Y \mid Z$, then the MSE obtained after regressing $Y$ on $X, Z$ is smaller than that obtained after regressing $Y$ on $Z$ only. This assumption doesn't always hold. Take the system where $X, Y, Z$ are one-dimensional variables and $Y\sim \mathcal{N}(Z, X)$. In other words, $Y$ is a Gaussian with mean $Z$ and variance $X$. Then, whether $X$ is known or not, the true regression line is simply $Y=Z$, and the MSE does not change. However, $Y$ clearly depends on $X$, even when $Z$ is given. This represents just one possible failure mode; we do not attempt to characterize all the cases in which nonlinear regression does not benefit from a conditionally dependent regressor and simply assume them away.

\subsection{Which test should I use?}
Our experimental results suggest that in general, FIT is the best choice. The only situation where an alternative should be considered is low-dimensional data where it i impossible to obtain more than 1000 samples. In such cases, CHSIC or KCIT are likely to outperform FIT.

RCIT and CCI do not work well in our evaluation. KCIPT is too slow to justify using it over FIT, KCIT or CHSIC.

\FloatBarrier
\bibliographystyle{plainnat} 
\bibliography{Zotero,bibliography}

\appendix
\section{Full Results}
\label{sec:full_res}
\begin{figure}[!h]
\includegraphics[width=1.\textwidth]{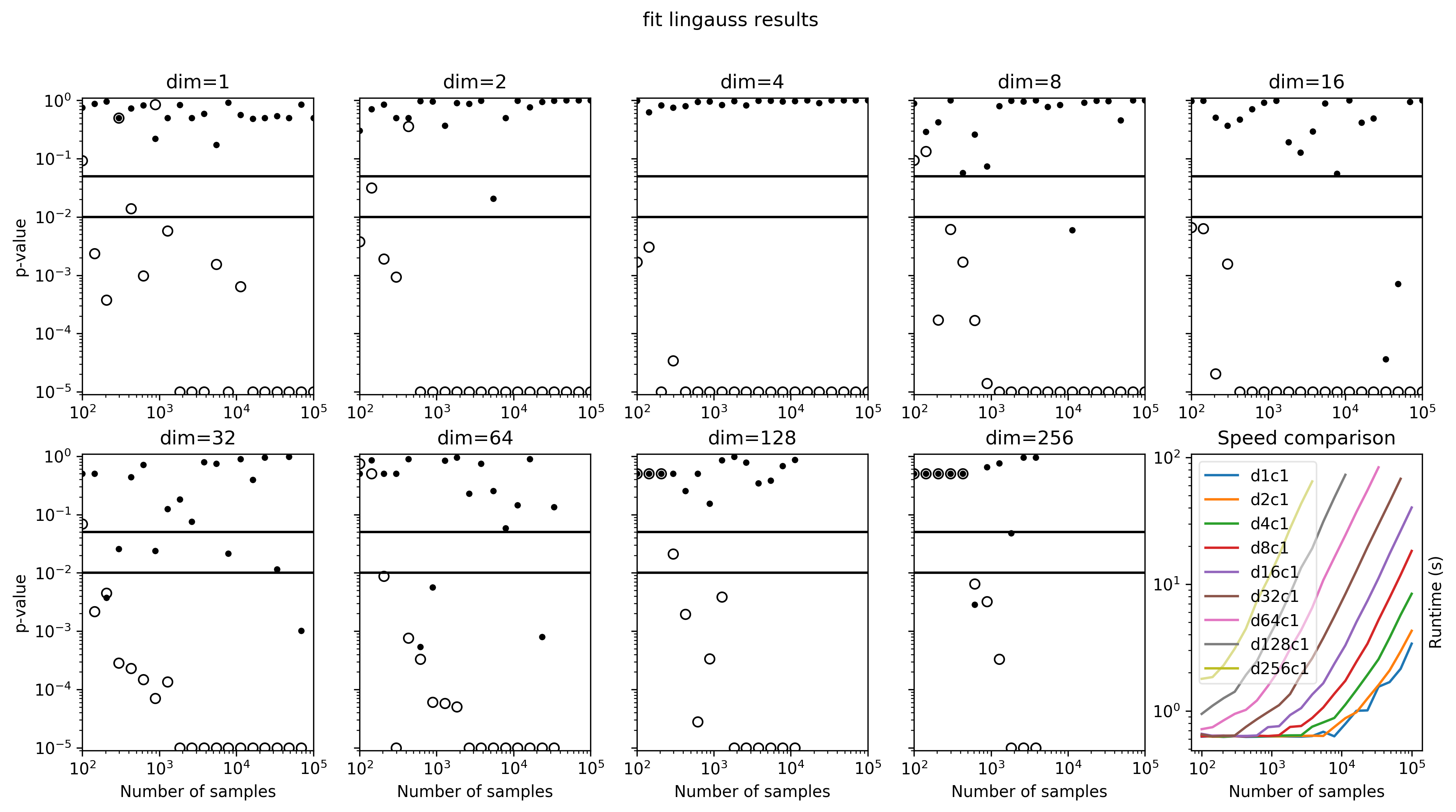}
\caption{}
\label{fig:result_chain_fit}
\end{figure}

\begin{figure}[!h]
\includegraphics[width=1.\textwidth]{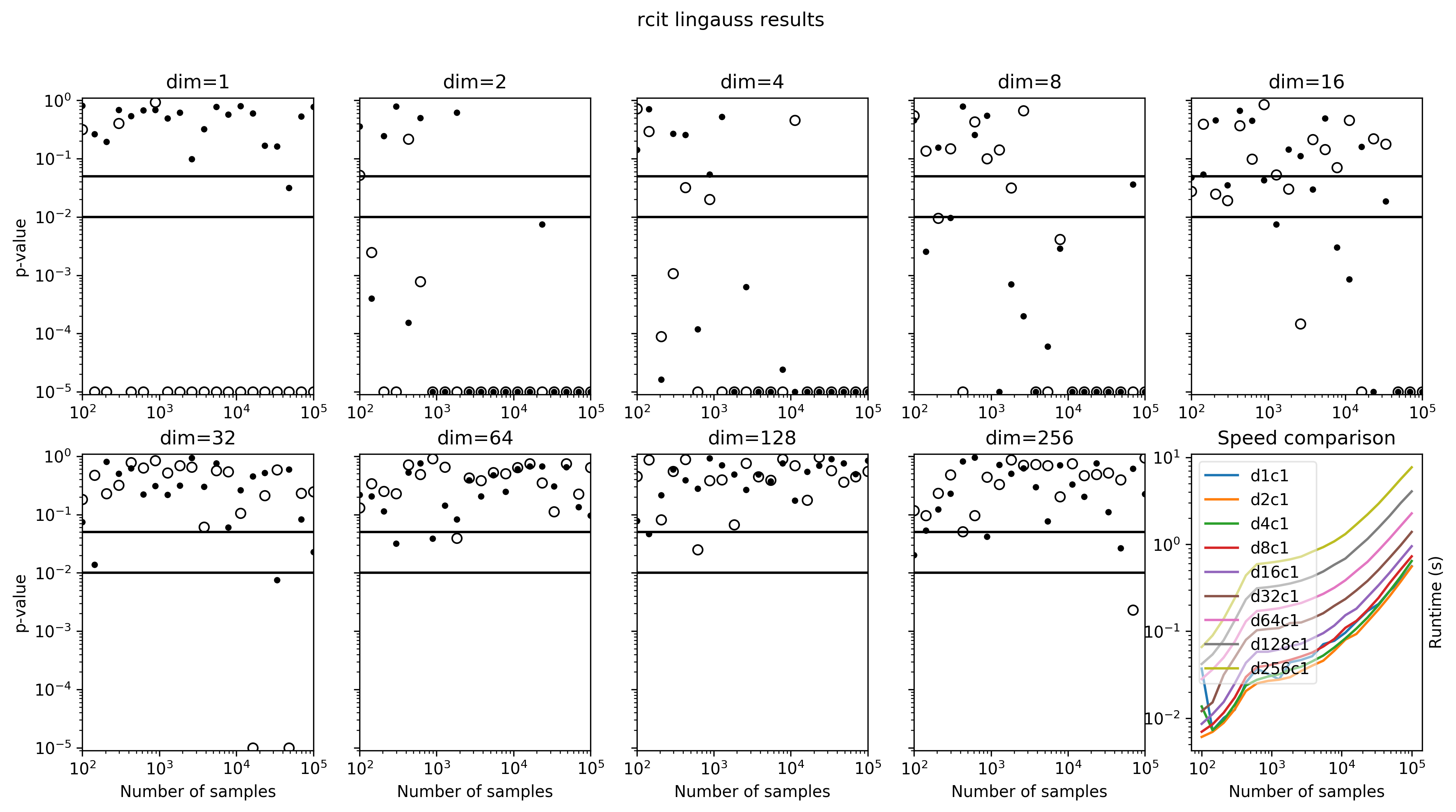}
\caption{}
\label{fig:result_chain_rcit}
\end{figure}

\begin{figure}[!h]
\includegraphics[width=1.\textwidth]{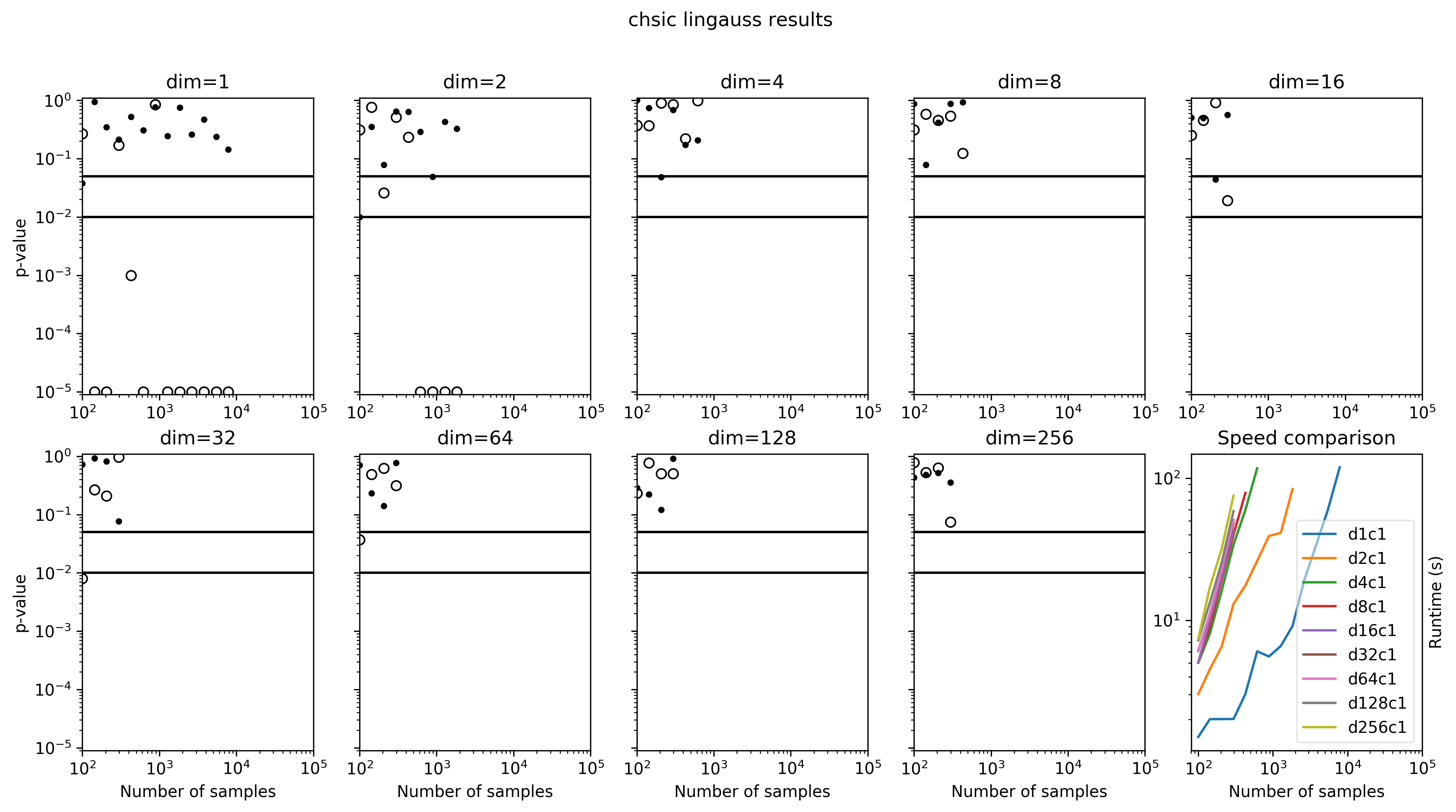}
\caption{}
\label{fig:result_chain_chsic}
\end{figure}

\begin{figure}[!h]
\includegraphics[width=1.\textwidth]{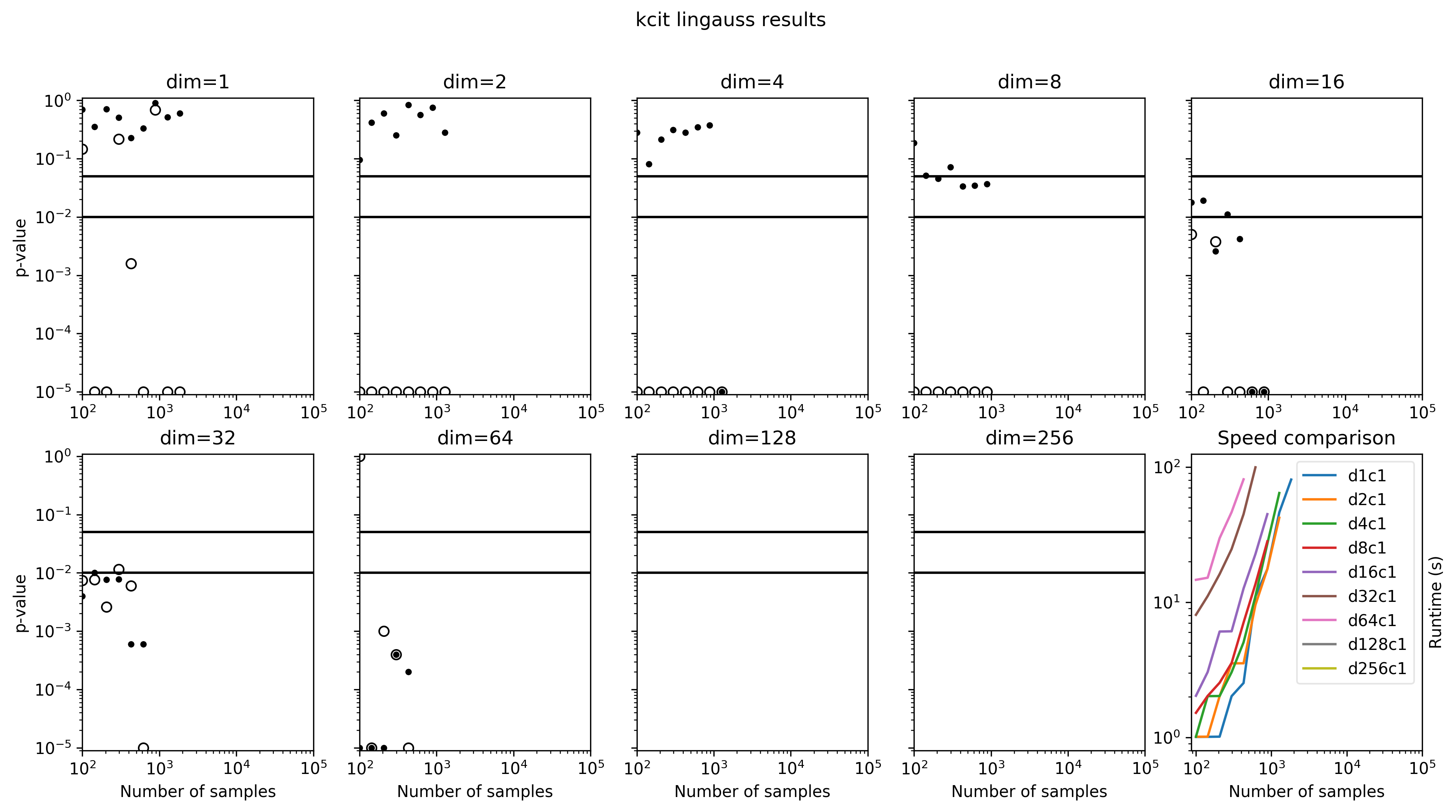}
\caption{}
\label{fig:result_chain_kcit}
\end{figure}

\begin{figure}[!h]
\includegraphics[width=1.\textwidth]{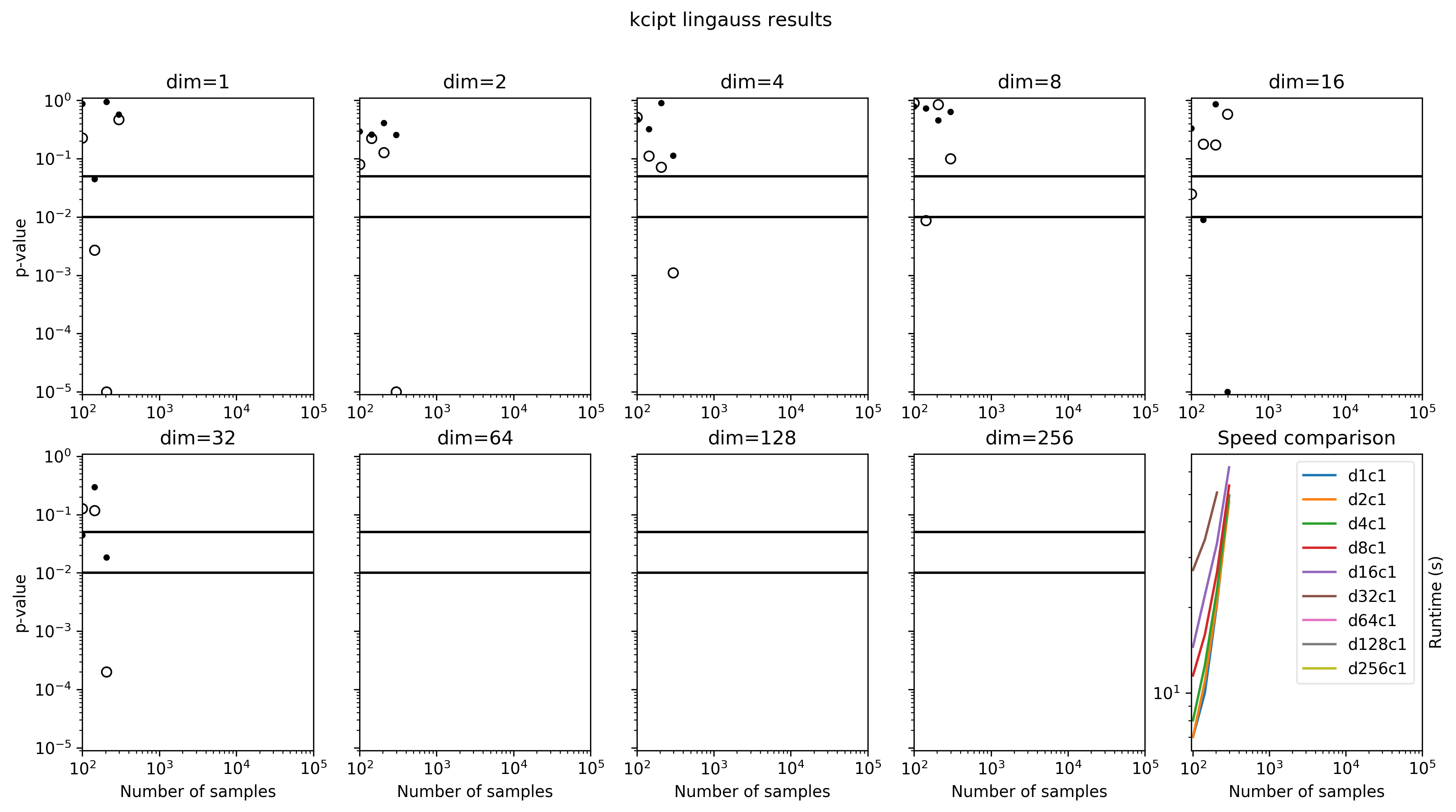}
\caption{}
\label{fig:result_chain_kcipt}
\end{figure}

\begin{figure}[!h]
\includegraphics[width=1.\textwidth]{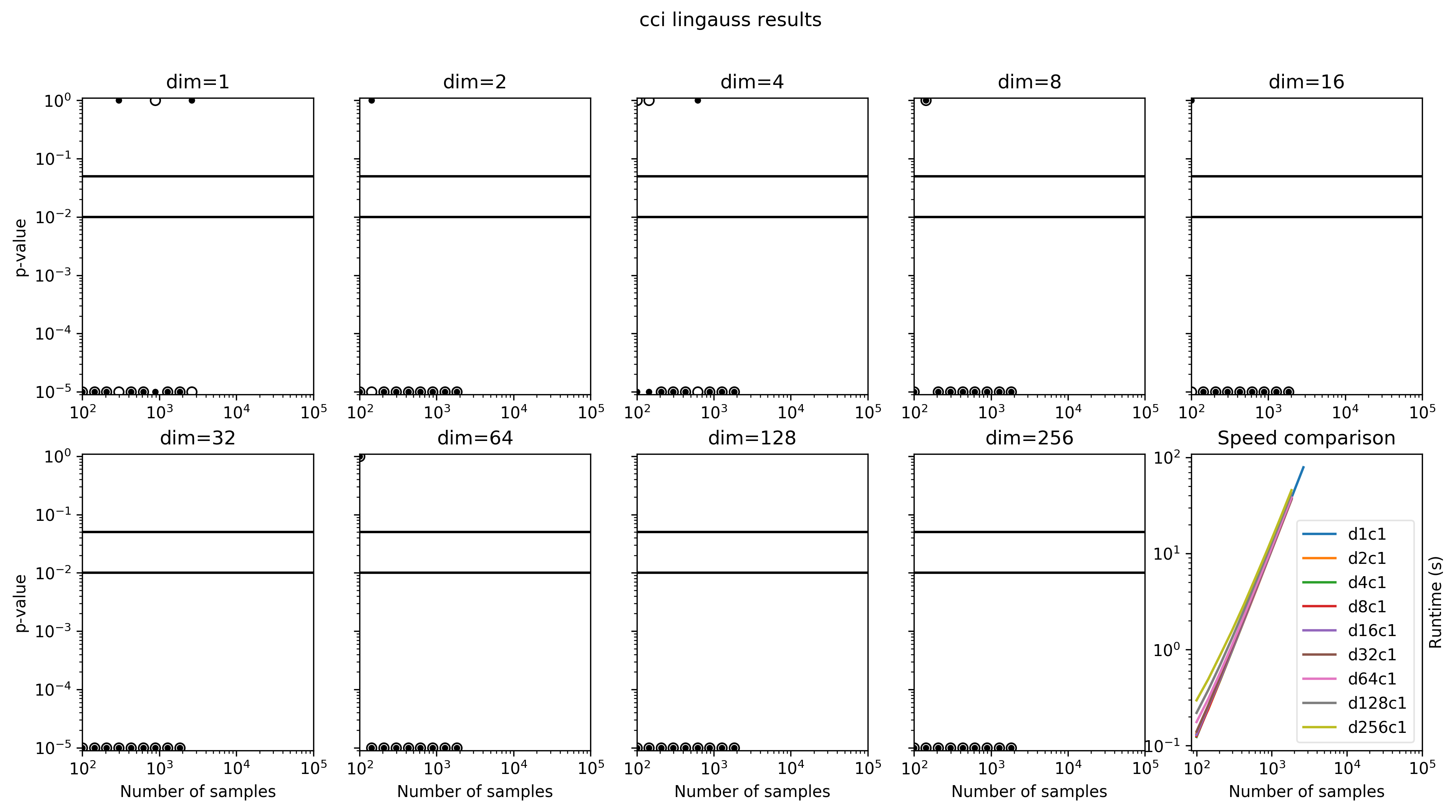}
\caption{}
\label{fig:result_chain_cci}
\end{figure}

\begin{figure}[!h]
\includegraphics[width=1.\textwidth]{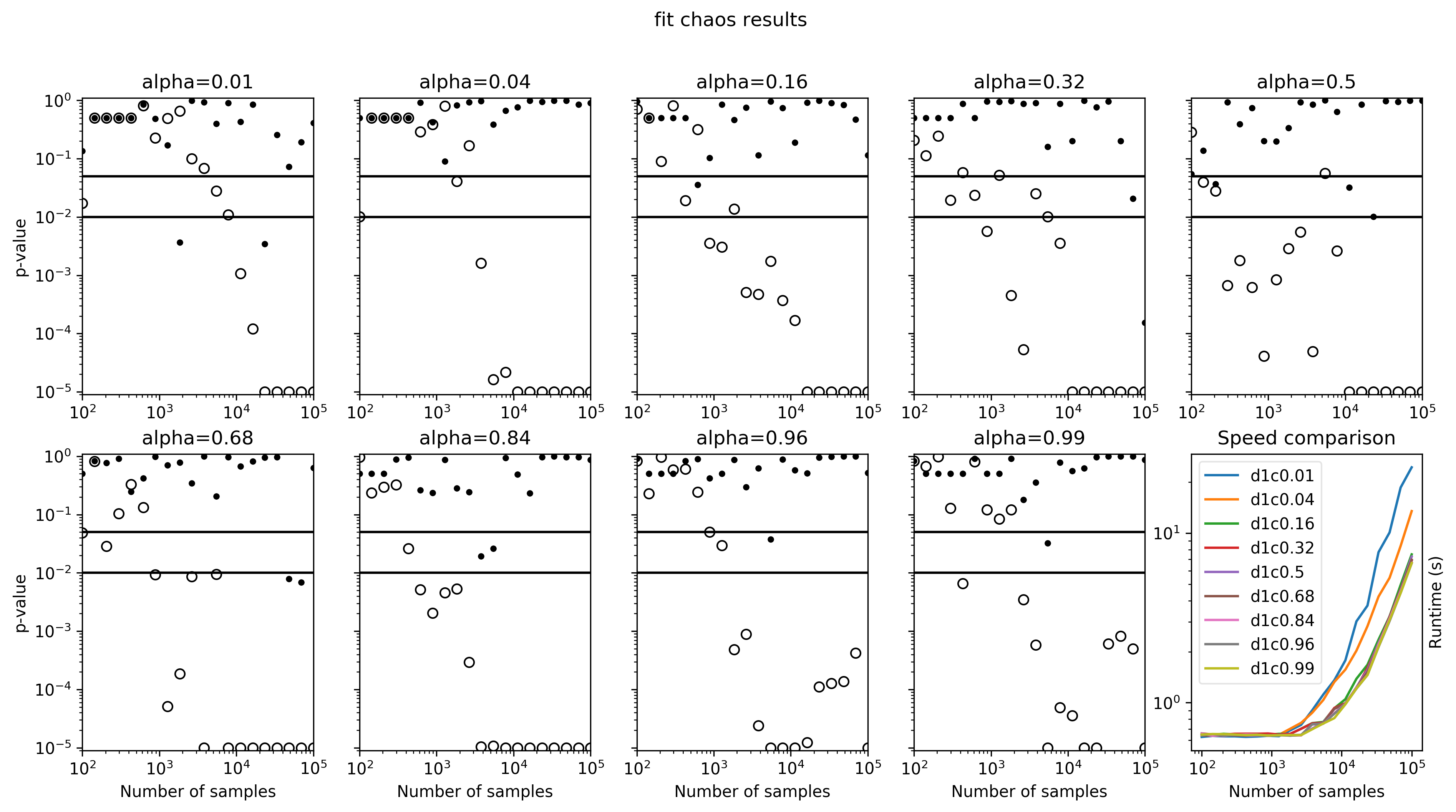}
\caption{}
\label{fig:result_chaos_fit}
\end{figure}

\begin{figure}[!h]
\includegraphics[width=1.\textwidth]{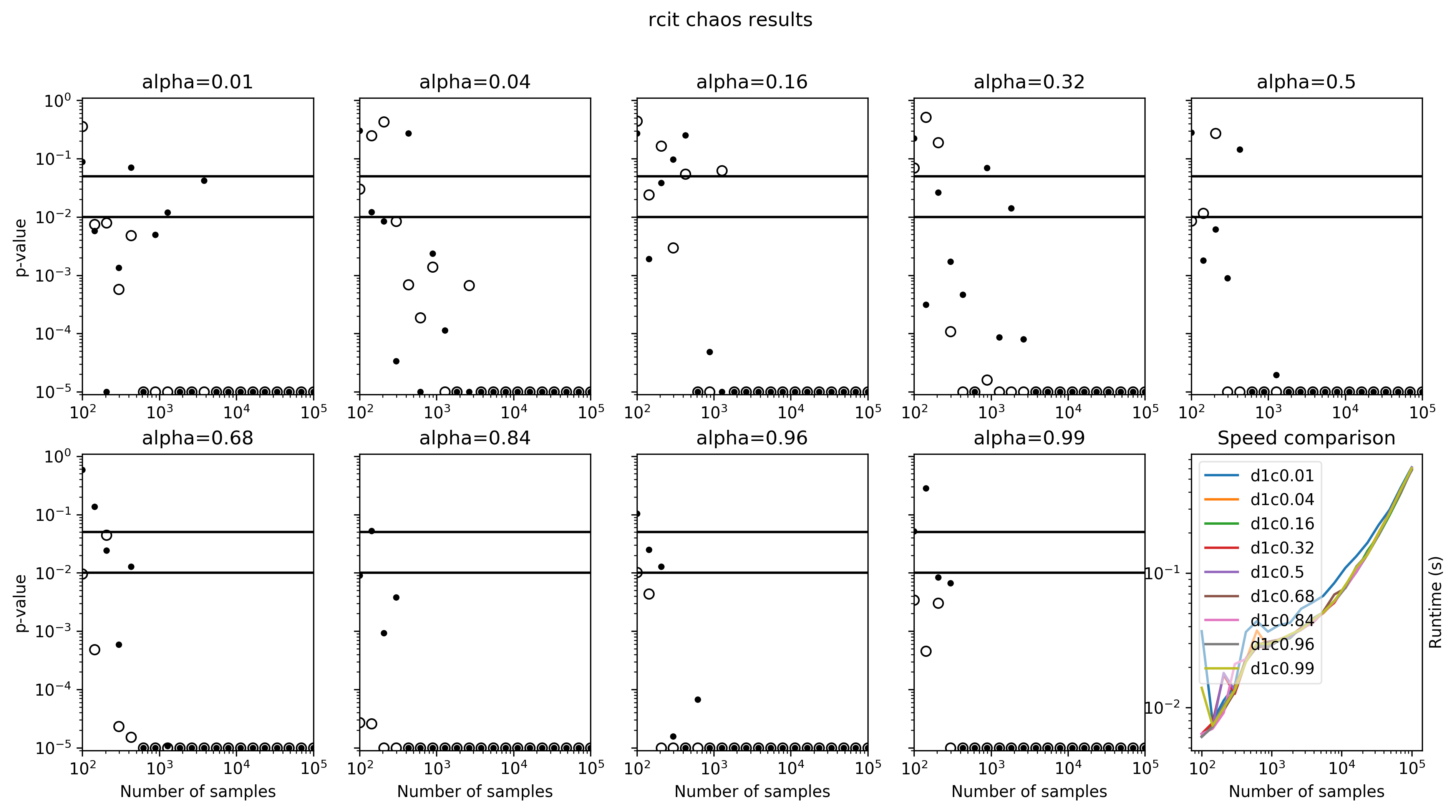}
\caption{}
\label{fig:result_chaos_rcit}
\end{figure}

\begin{figure}[!h]
\includegraphics[width=1.\textwidth]{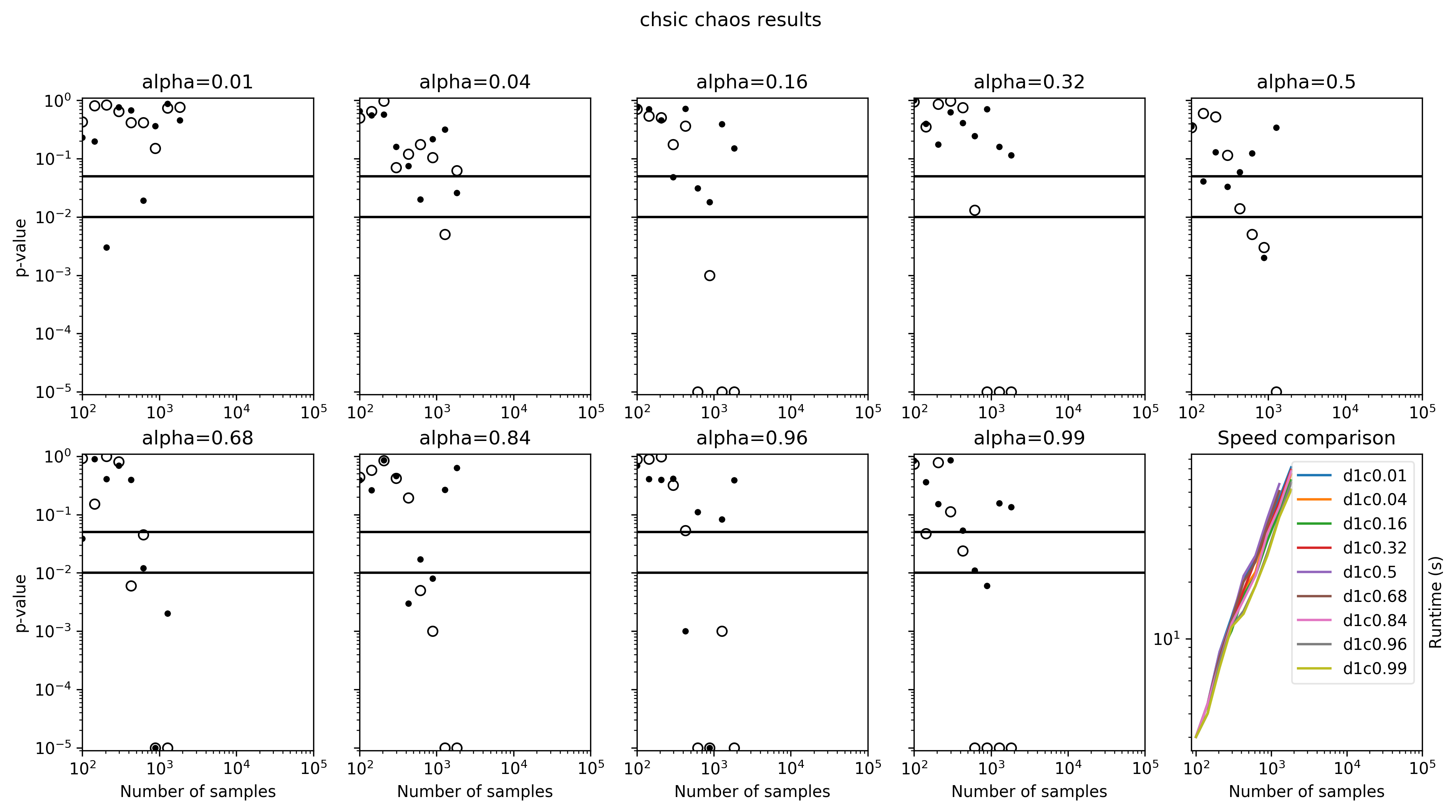}
\caption{}
\label{fig:result_chaos_chsic}
\end{figure}

\begin{figure}[!h]
\includegraphics[width=1.\textwidth]{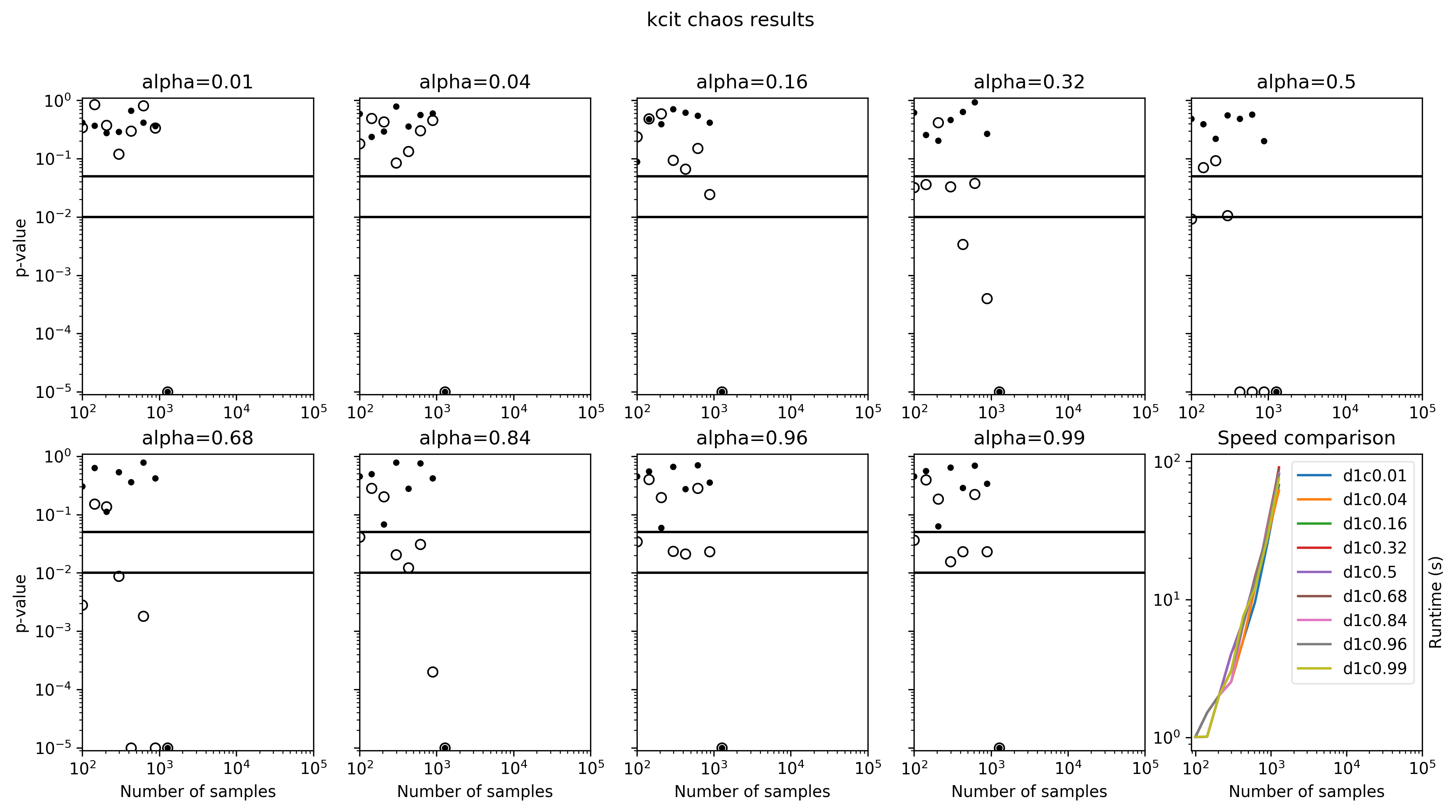}
\caption{}
\label{fig:result_chaos_kcit}
\end{figure}

\begin{figure}[!h]
\includegraphics[width=1.\textwidth]{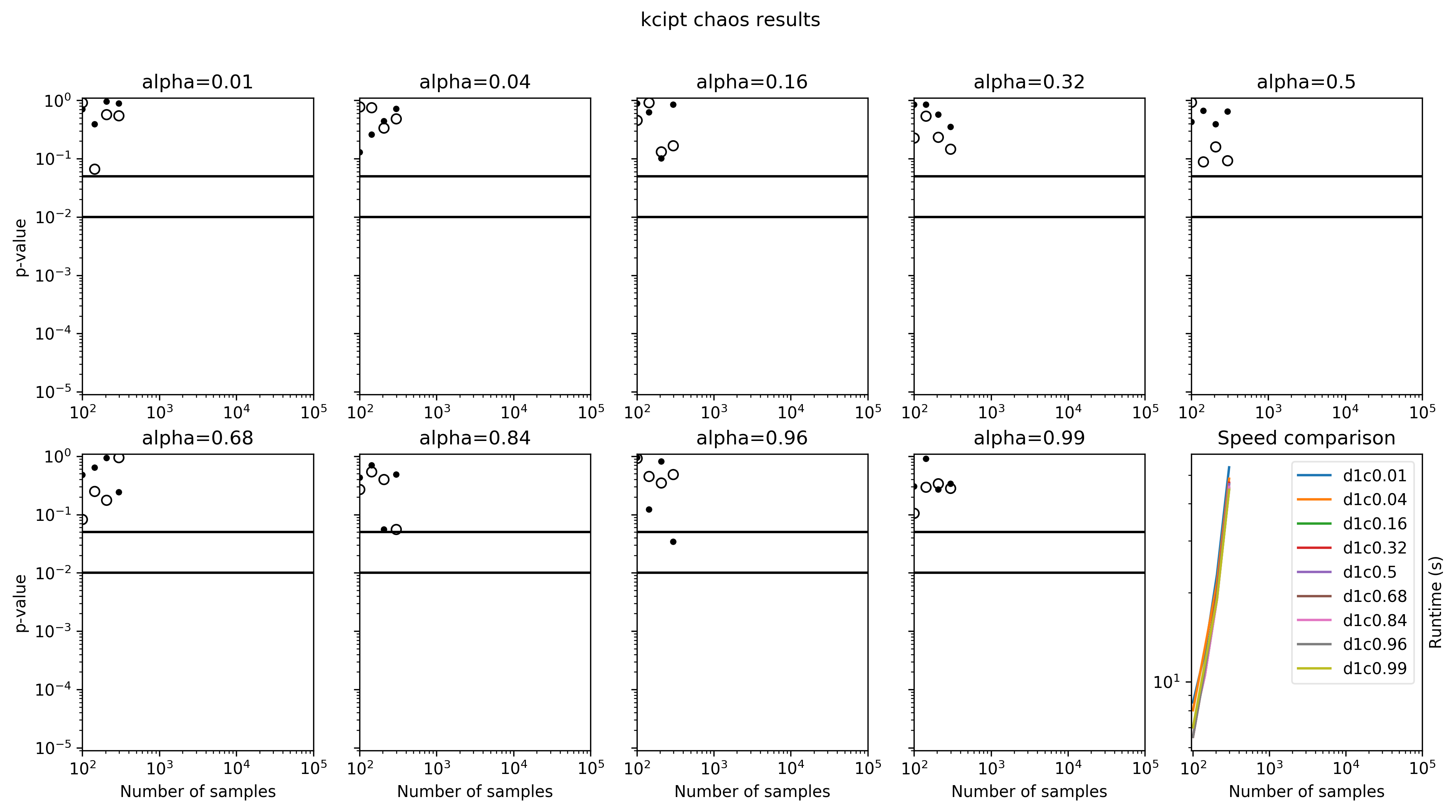}
\caption{}
\label{fig:result_chaos_kcipt}
\end{figure}

\begin{figure}[!h]
\includegraphics[width=1.\textwidth]{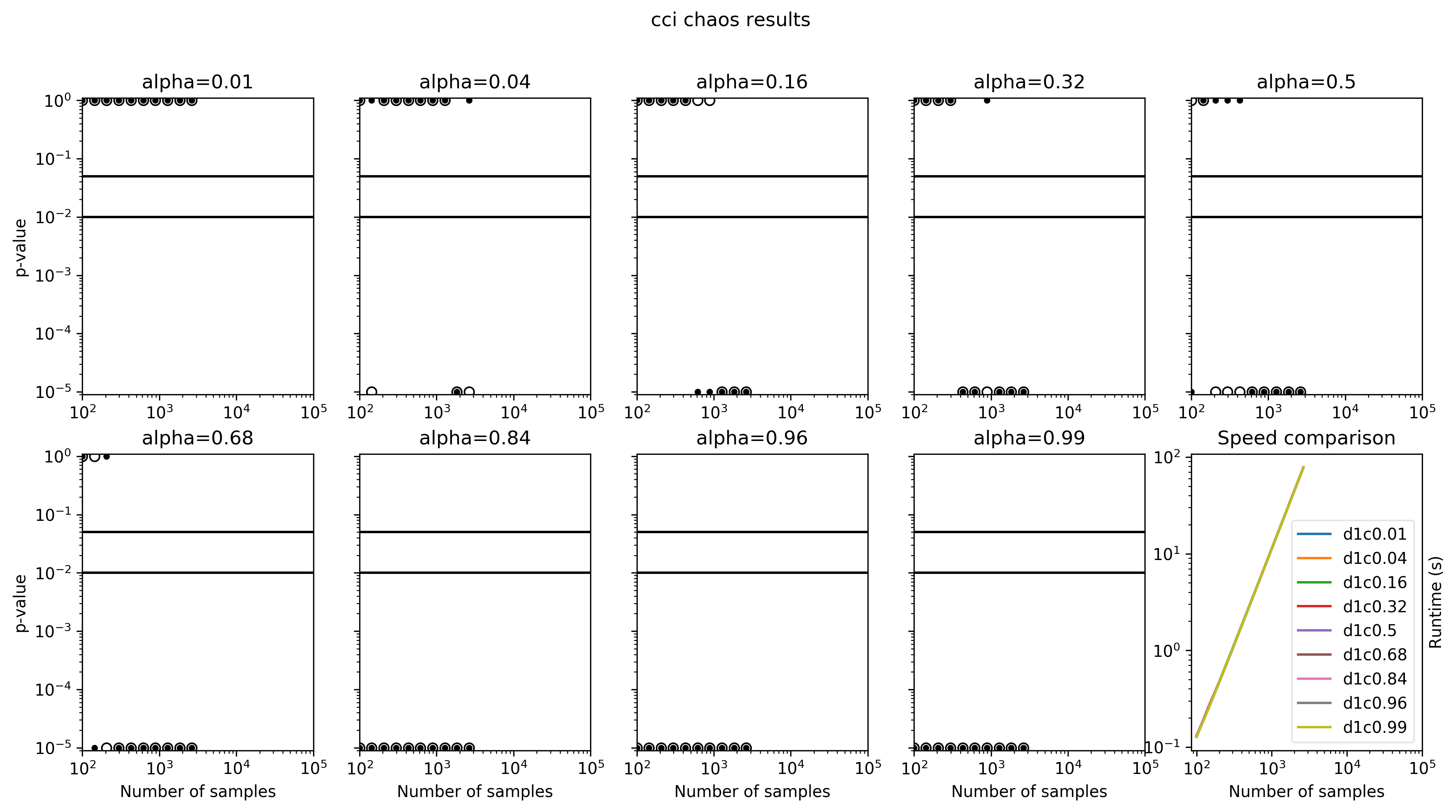}
\caption{}
\label{fig:result_chaos_cci}
\end{figure}

\begin{figure}[!h]
\includegraphics[width=1.\textwidth]{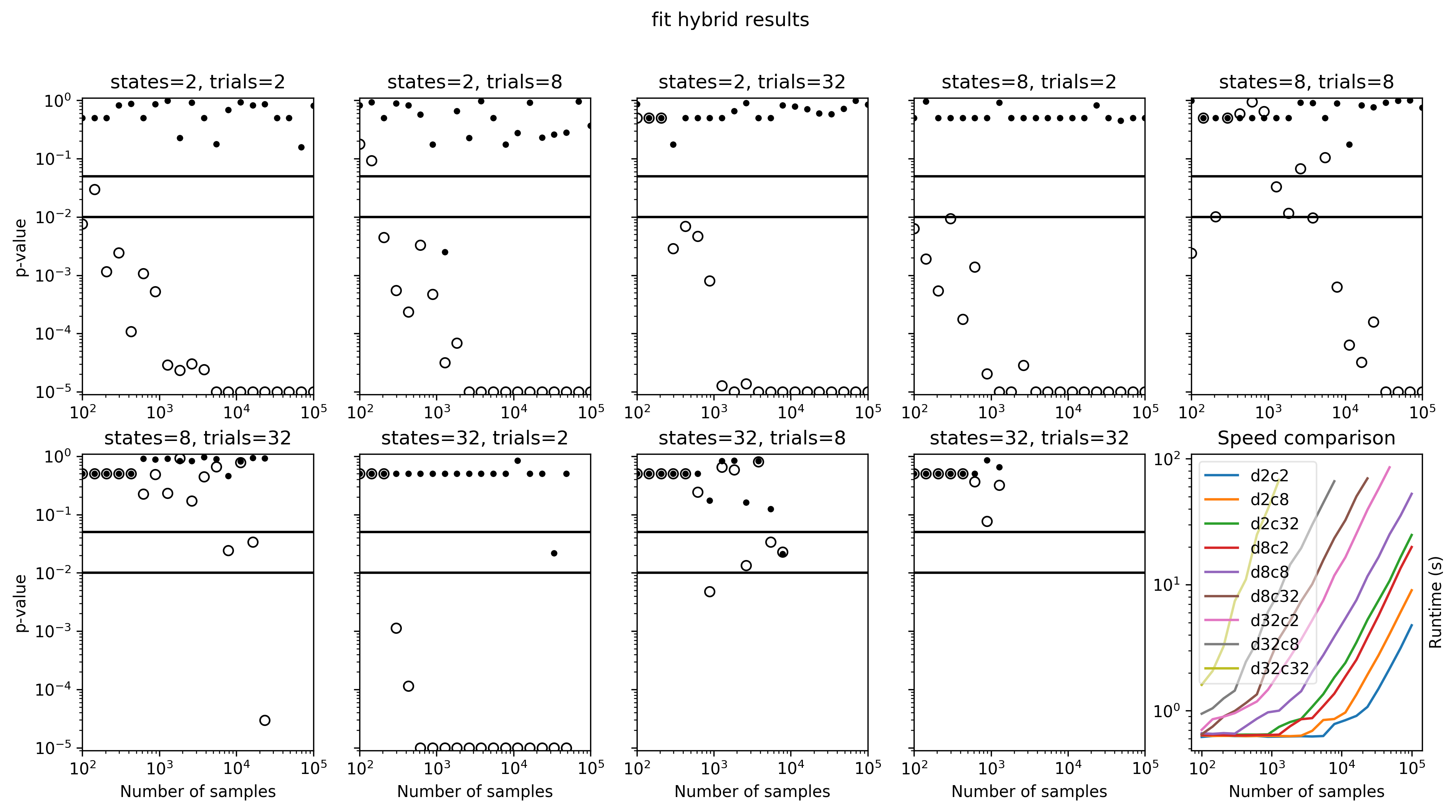}
\caption{}
\label{fig:result_discrete_fit}
\end{figure}

\begin{figure}[!h]
\includegraphics[width=1.\textwidth]{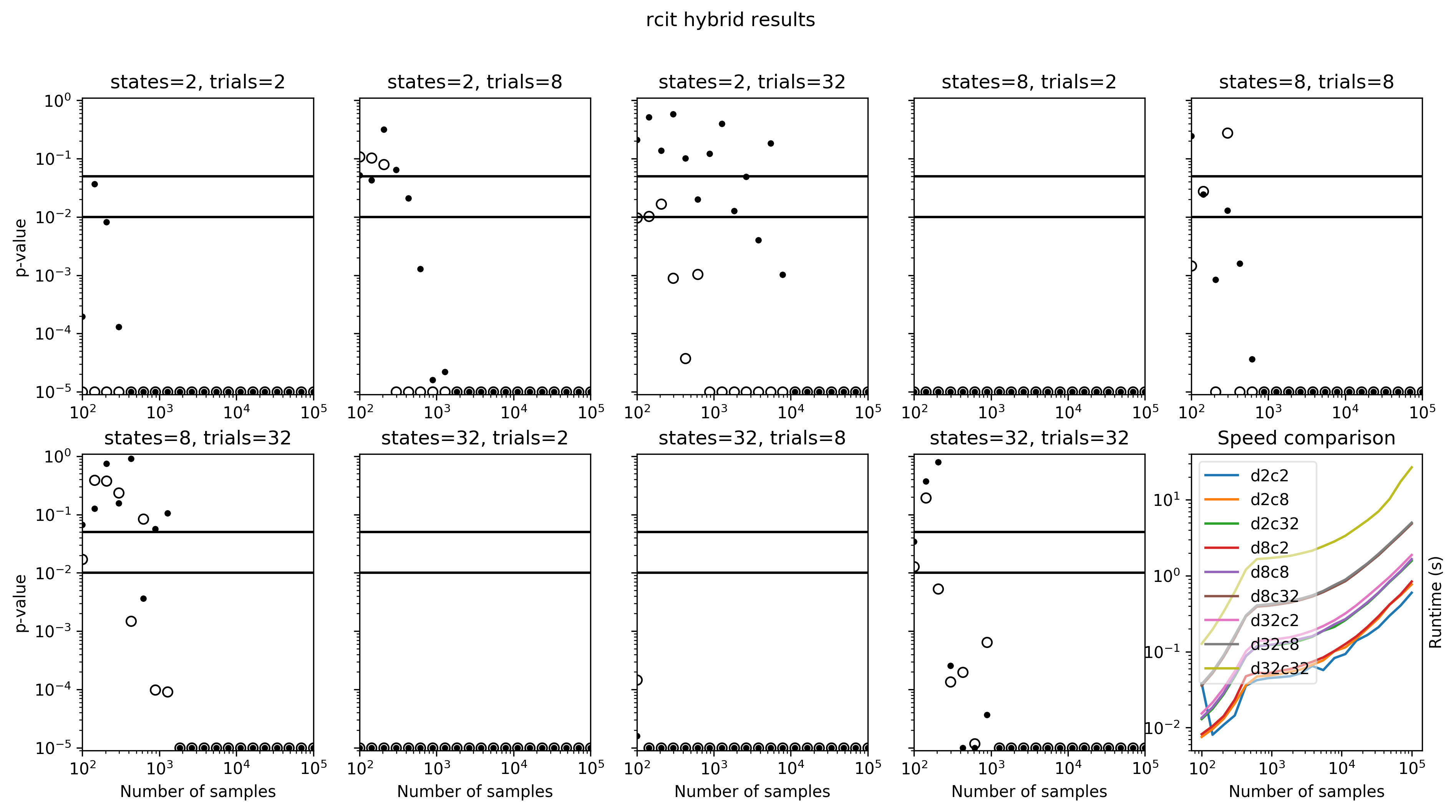}
\caption{}
\label{fig:result_discrete_rcit}
\end{figure}

\begin{figure}[!h]
\includegraphics[width=1.\textwidth]{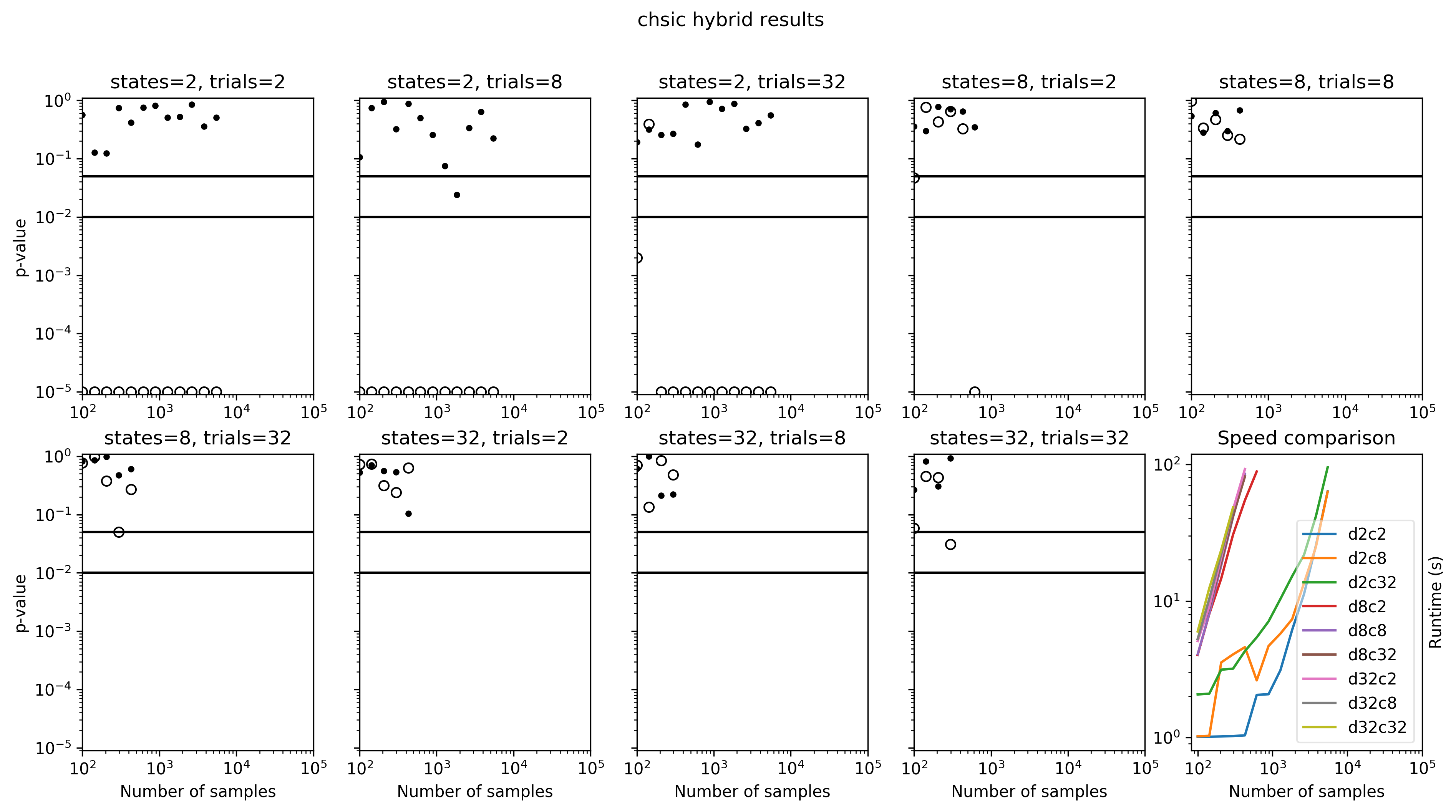}
\caption{}
\label{fig:result_discrete_chsic}
\end{figure}

\begin{figure}[!h]
\includegraphics[width=1.\textwidth]{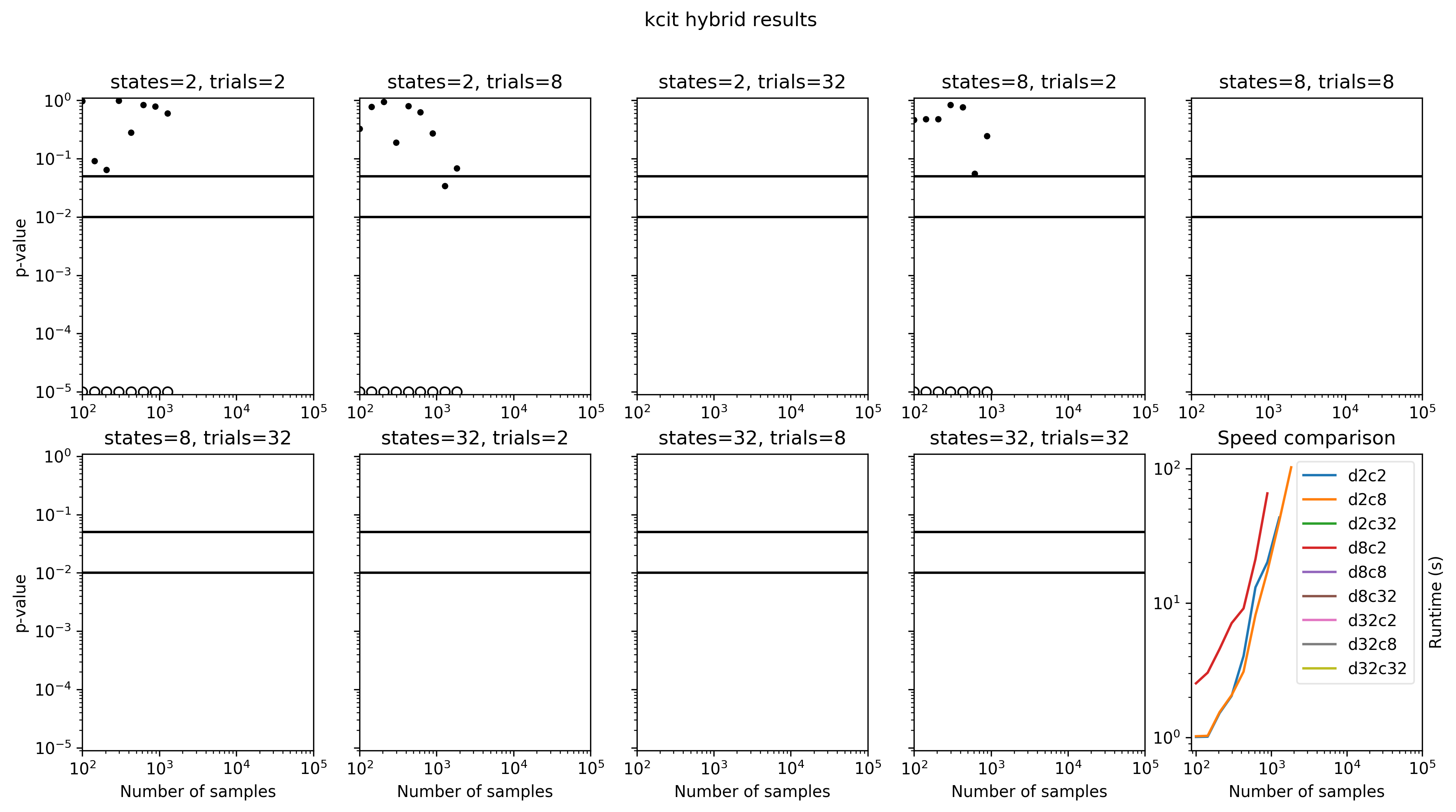}
\caption{}
\label{fig:result_discrete_kcit}
\end{figure}

\begin{figure}[!h]
\includegraphics[width=1.\textwidth]{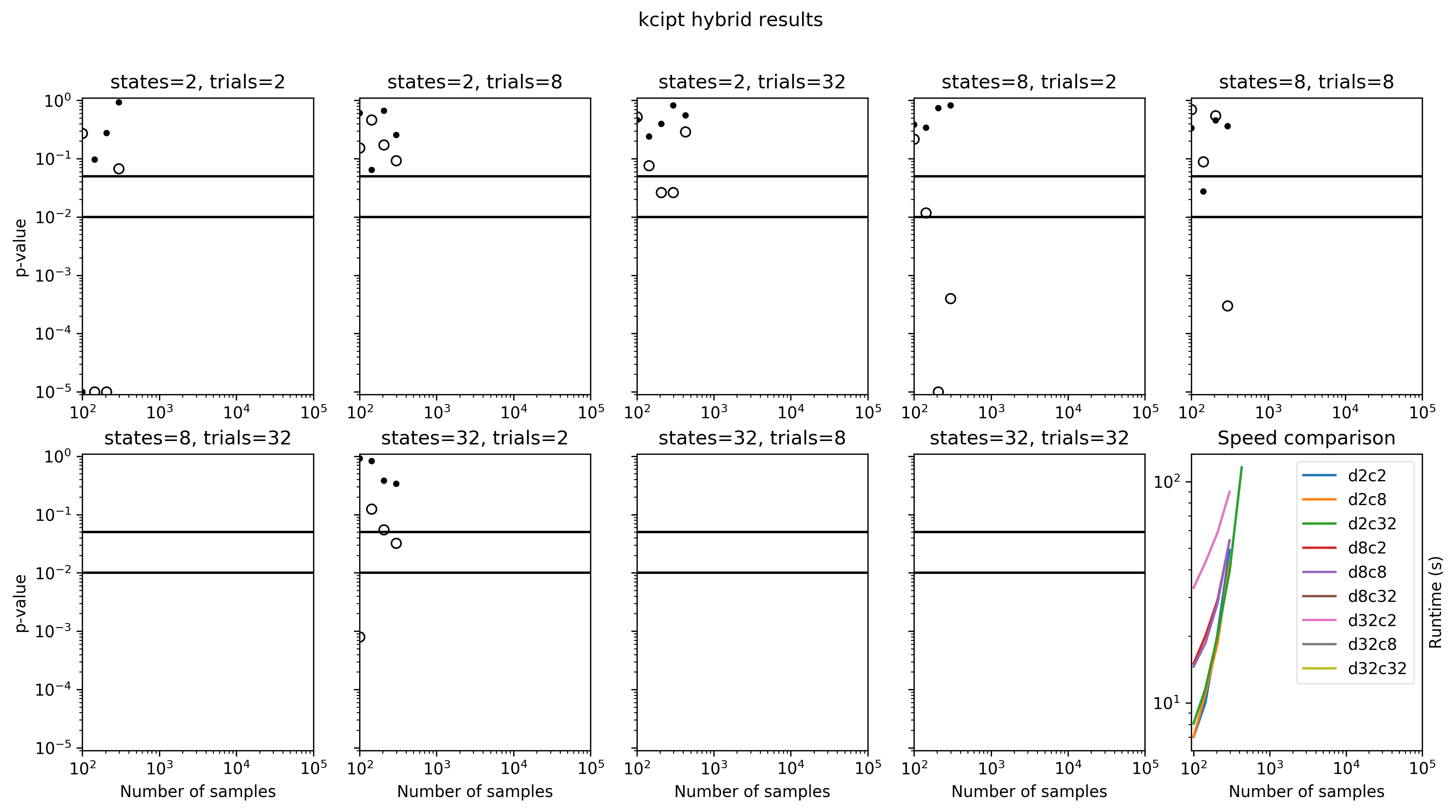}
\caption{}
\label{fig:result_discrete_kcipt}
\end{figure}

\begin{figure}[!h]
\includegraphics[width=1.\textwidth]{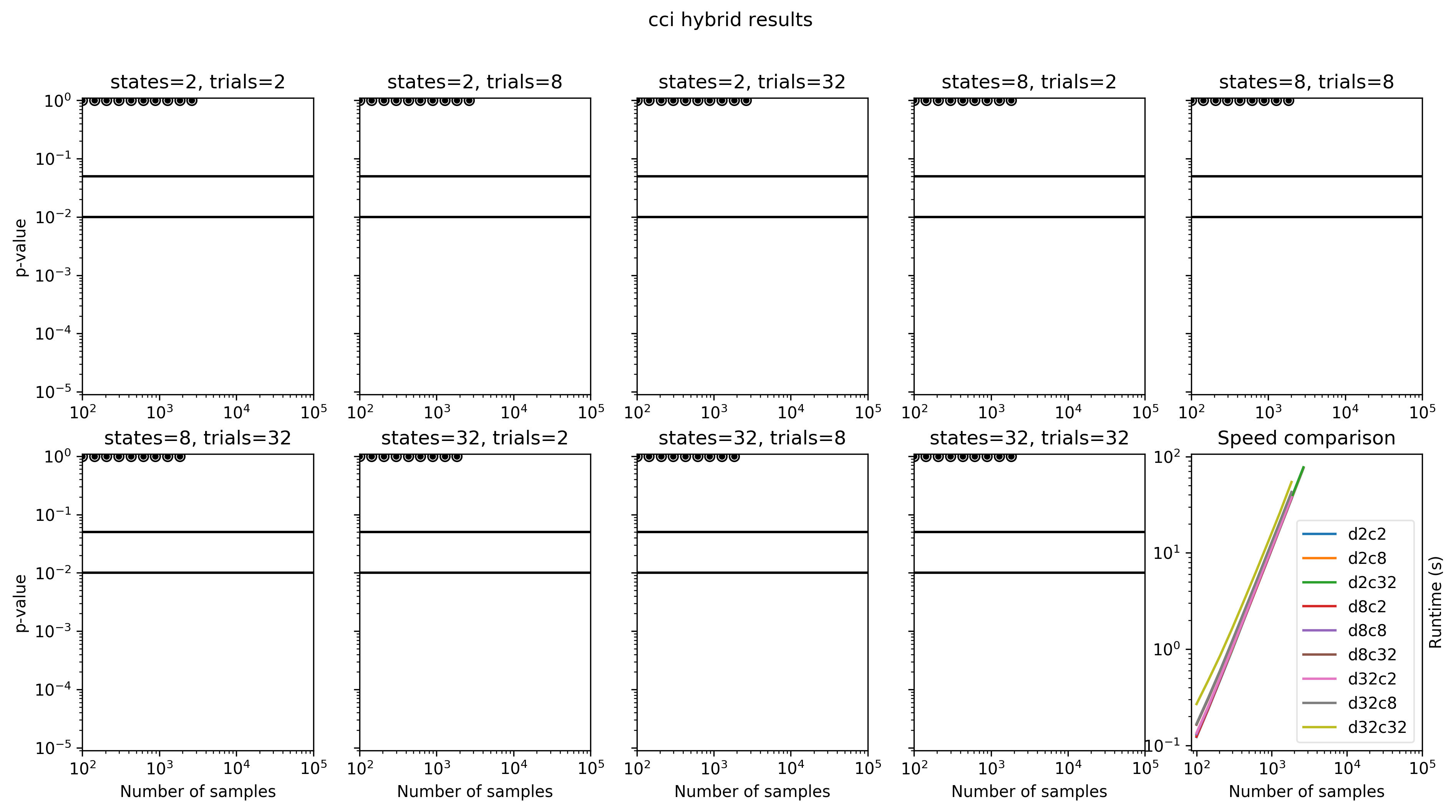}
\caption{}
\label{fig:result_discrete_cci}
\end{figure}

\begin{figure}[!h]
\includegraphics[width=1.\textwidth]{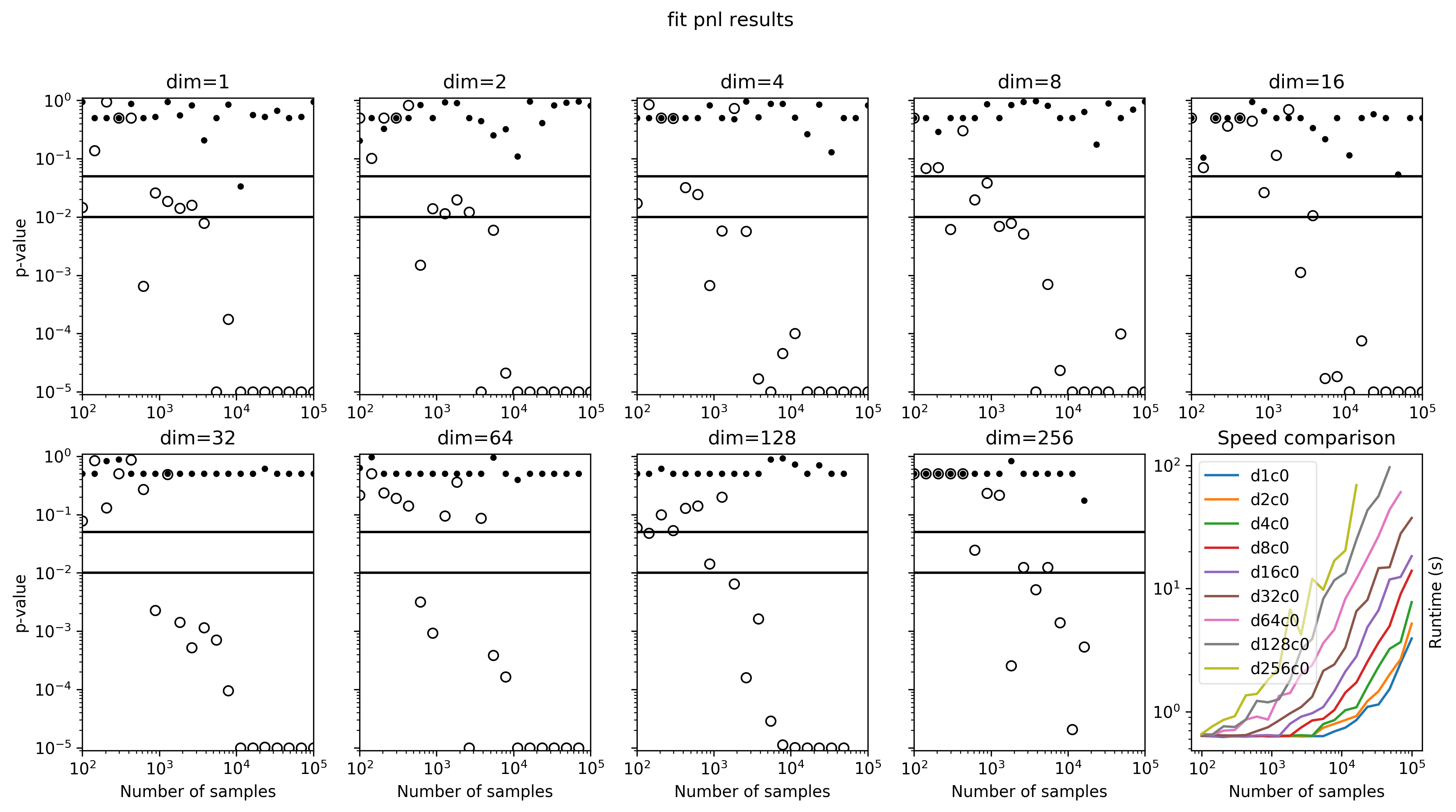}
\caption{}
\label{fig:result_pnl_fit}
\end{figure}

\begin{figure}
\includegraphics[width=1.\textwidth]{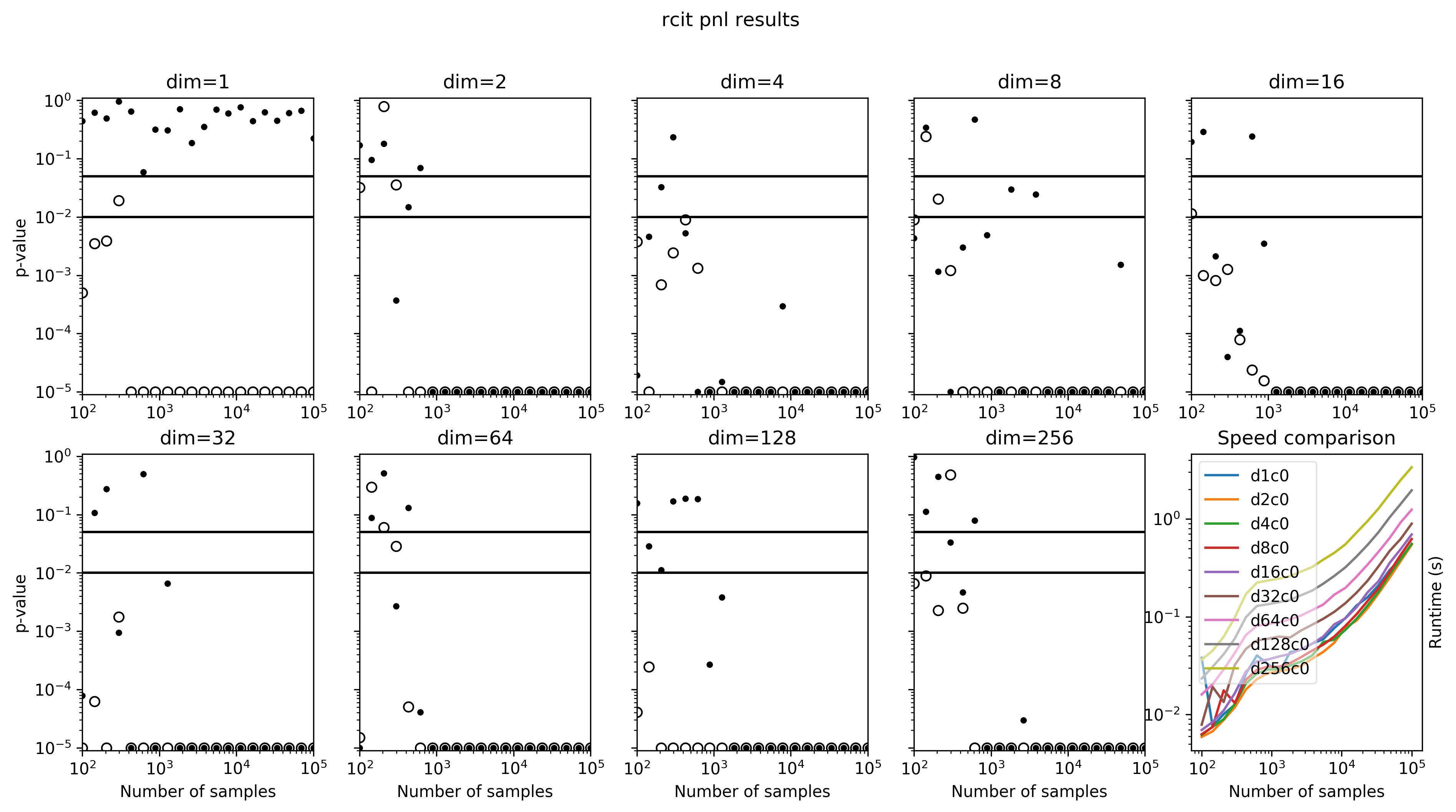}
\caption{}
\label{fig:result_pnl_rcit}
\end{figure}

\begin{figure}[!h]
\includegraphics[width=1.\textwidth]{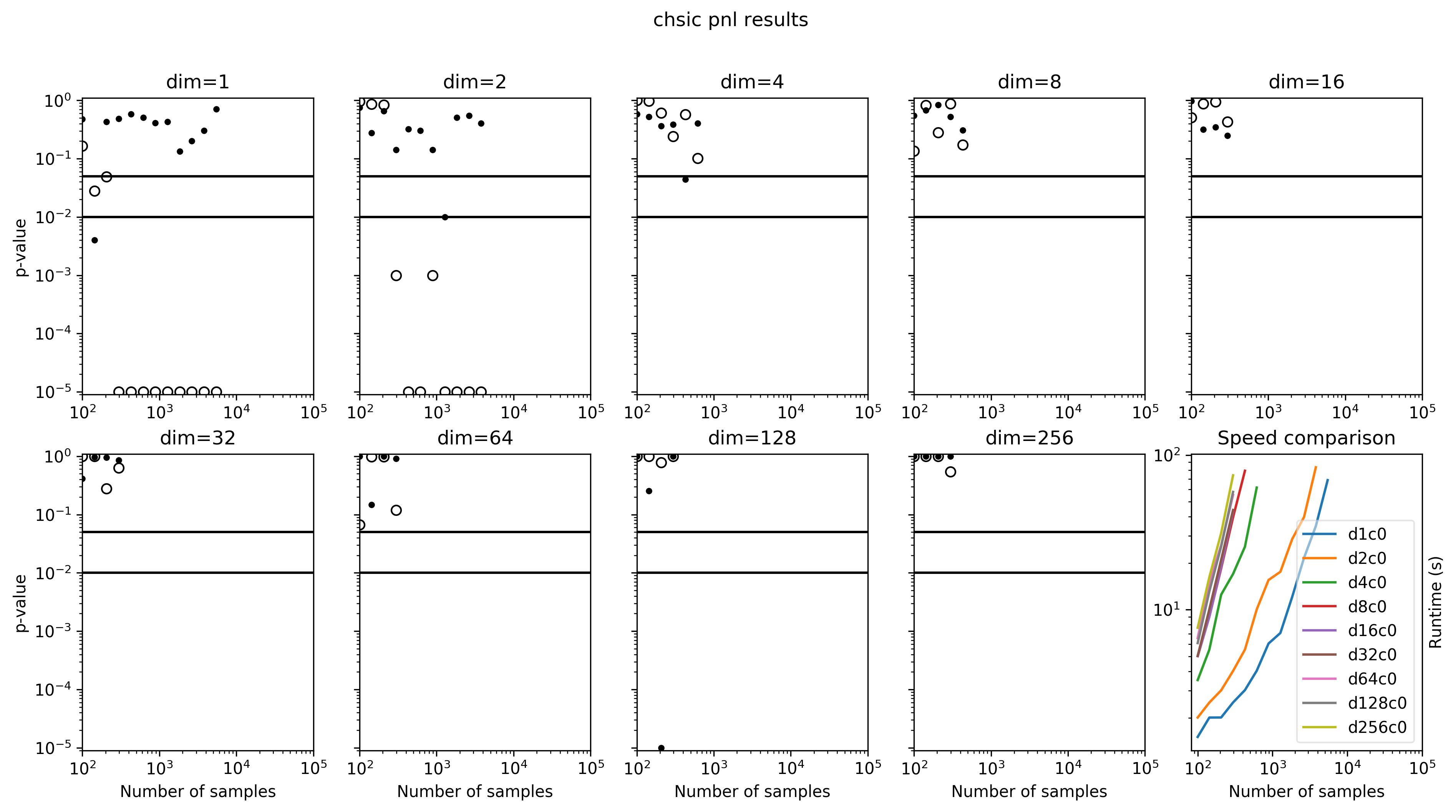}
\caption{}
\label{fig:result_pnl_chsic}
\end{figure}

\begin{figure}[!h]
\includegraphics[width=1.\textwidth]{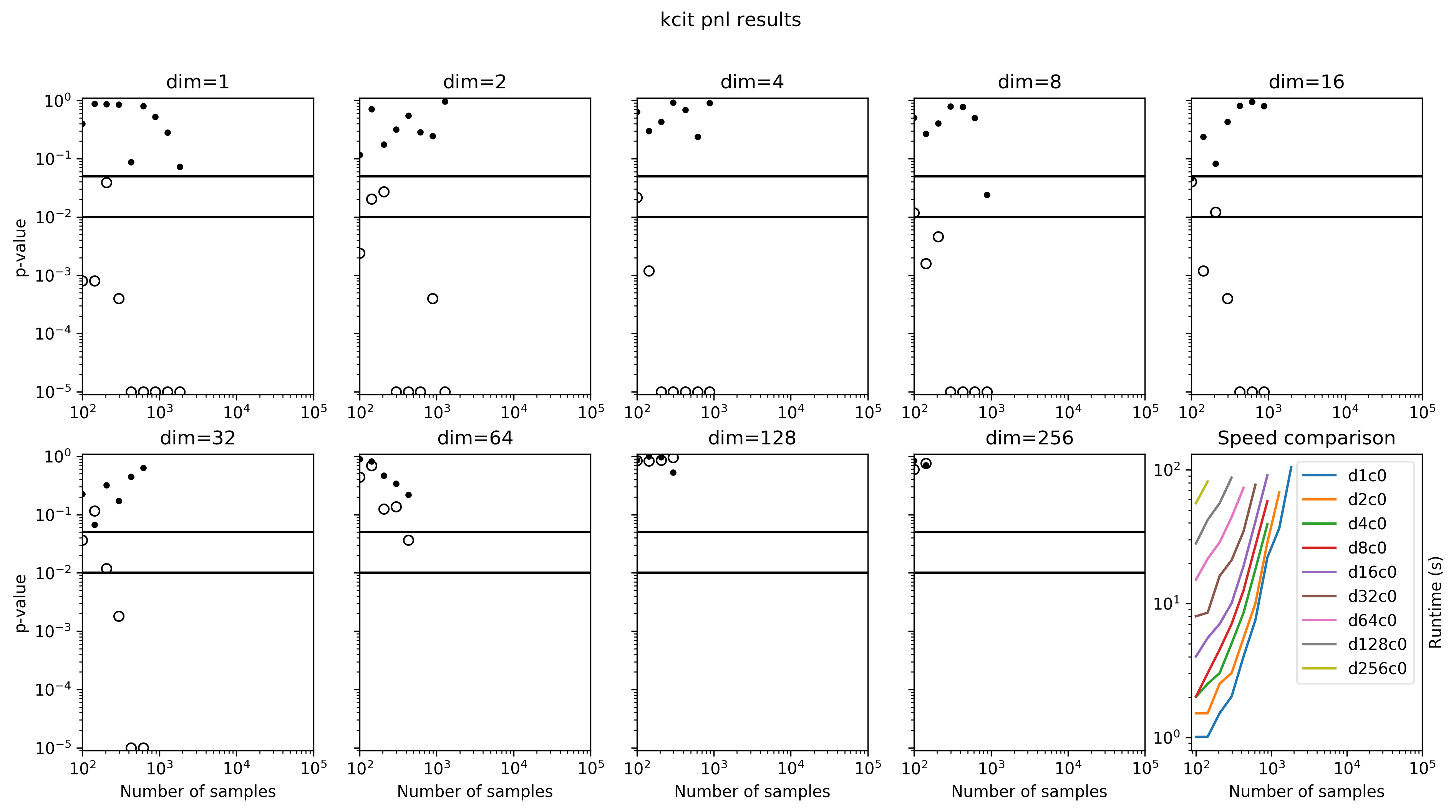}
\caption{}
\label{fig:result_pnl_kcit}
\end{figure}

\begin{figure}[!h]
\includegraphics[width=1.\textwidth]{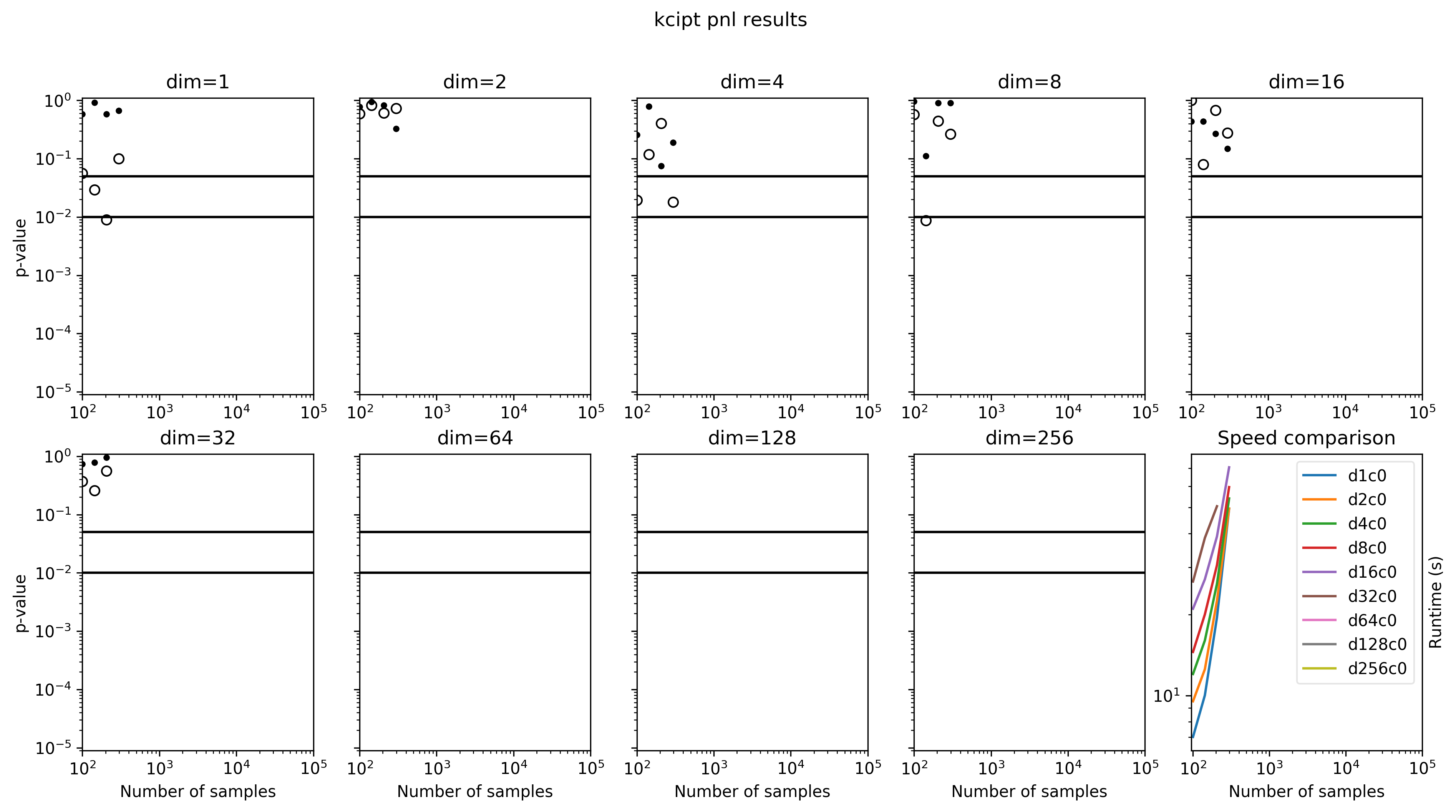}
\caption{}
\label{fig:result_pnl_kcipt}
\end{figure}

\begin{figure}[!h]
\includegraphics[width=1.\textwidth]{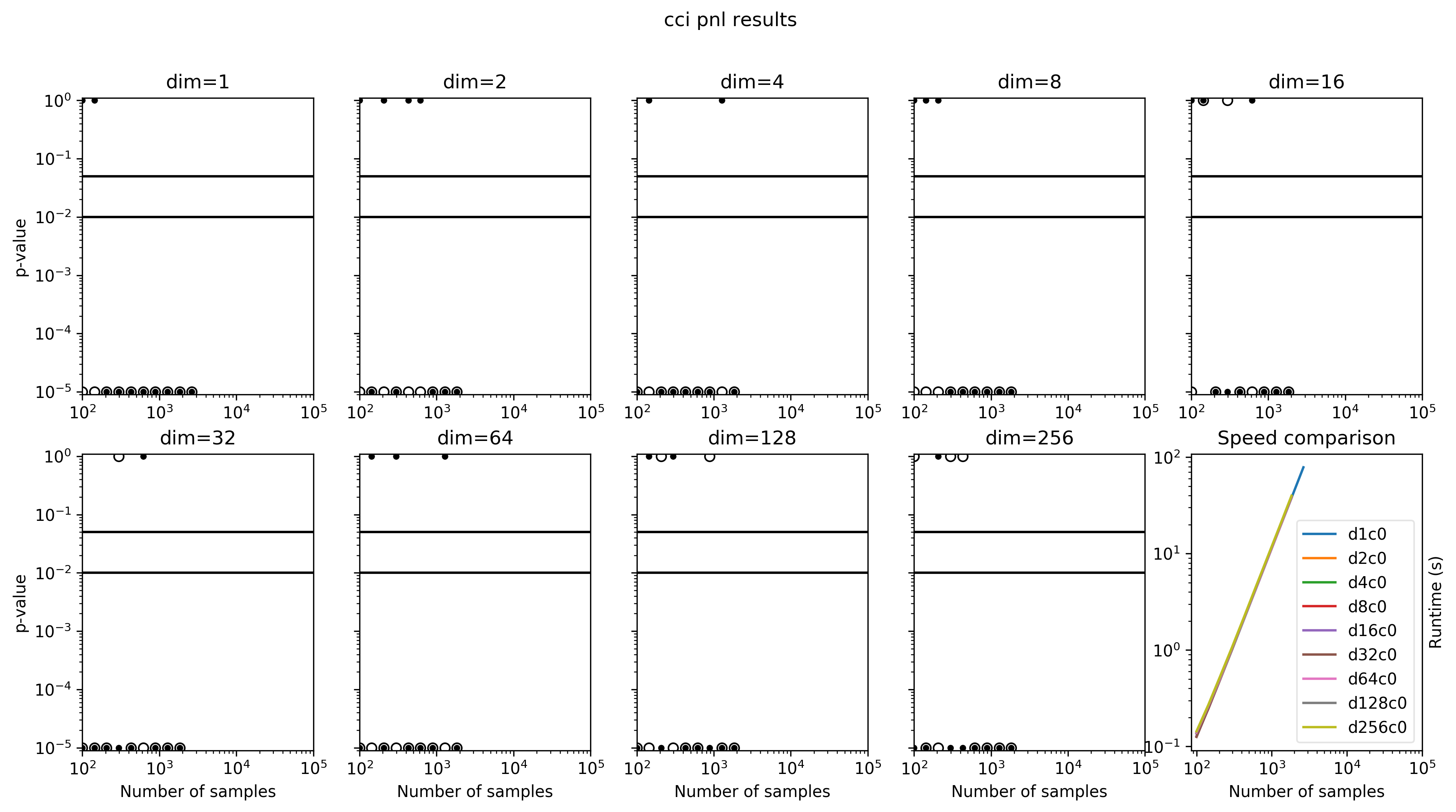}
\caption{}
\label{fig:result_pnl_cci}
\end{figure}

Below, we present the full results plots for all the datasets and methods. Fig.~\ref{fig:minires} compresses (not losslessly!) information presented here. In addition, the full results support the conclusions of Sec.~\ref{sec:discussion}.  Figures~\ref{fig:result_chain_fit}-~\ref{fig:result_pnl_cci} show plots of $p$-values obtained by each algorithm as the dataset and number of samples varies.

\subsection{\textsc{lingauss} datasets}

\FloatBarrier
\subsection{\textsc{chaos} datasets}

\FloatBarrier
\subsection{\textsc{hybrid} datasets}

\FloatBarrier
\subsection{\textsc{pnl} datasets}

\clearpage
\end{document}